\documentclass[twoside,11pt]{article}

\usepackage{microtype}
\usepackage{graphicx}
\usepackage{booktabs} 
\usepackage[lofdepth,lotdepth]{subfig}
\usepackage{jmlr2e}

\usepackage{algorithm}
\usepackage{algorithmic}

\usepackage{amsmath}
\usepackage{amssymb}
\usepackage{bm}
\usepackage{pdfpages}
\usepackage{multirow}
\usepackage{soul}

\newcommand\blfootnote[1]{%
  \begingroup
  \renewcommand\thefootnote{}\footnote{#1}%
  \addtocounter{footnote}{-1}%
  \endgroup
}

\ShortHeadings{Intrinsic Exploration as Multi-Objective RL}{Morere and Ramos}
\firstpageno{1}

\begin{document}

\title{Intrinsic Exploration as Multi-Objective RL}

\author{
  \name Philippe Morere \email philippe.morere@sydney.edu.au\\
  \addr The University of Sydney,\\
  Sydney, Australia\\
  \AND
  \name Fabio Ramos \email fabio.ramos@sydney.edu.au\\
  \addr The University of Sydney \& NVIDIA,\\
  Sydney, Australia}


\maketitle

\begin{abstract}
Intrinsic motivation enables reinforcement learning (RL) agents to explore when rewards are very sparse, where traditional exploration heuristics such as Boltzmann or $\epsilon$-greedy would typically fail.
However, intrinsic exploration is generally handled in an ad-hoc manner, where exploration is not treated as a core objective of the learning process; this weak formulation leads to sub-optimal exploration performance.
To overcome this problem, we propose a framework based on multi-objective RL where both exploration and exploitation are being optimized as separate objectives.
This formulation brings the balance between exploration and exploitation at a policy level, resulting in advantages over traditional methods.
This also allows for controlling exploration while learning, at no extra cost. Such strategies achieve a degree of control over agent exploration that was previously unattainable with classic or intrinsic rewards.
We demonstrate scalability to continuous state-action spaces by presenting a method (EMU-Q) based on our framework, guiding exploration towards regions of higher value-function uncertainty.
EMU-Q is experimentally shown to outperform classic exploration techniques and other intrinsic RL methods on a continuous control benchmark and on a robotic manipulator.
\end{abstract}

\begin{keywords}
Reinforcement Learning, Robotics, Exploration, Online Learning, Multi-Objective
\end{keywords}

\section{Introduction}
In\blfootnote{This paper extends previous work published as \cite{morere2018bayesian}.} Reinforcement Learning (RL), data-efficiency and learning speed are paramount.
Indeed, when interacting with robots, humans, or the real world, data can be extremely scarce and expensive to collect. Improving data-efficiency is of the utmost importance to apply RL to interesting and realistic applications.
Learning from few data is relatively easier to achieve when rewards are dense, as these can be used to guide exploration.
In most realistic problems however, defining dense reward functions is non-trivial, requires expert knowledge and much fine-tuning. In some cases (eg. when dealing with humans), definitions for dense rewards are unclear and remain an open problem. This greatly hinders the applicability of RL to many interesting problems.

It appears more natural to reward robots only when reaching a goal, termed \emph{goal-only rewards}, which becomes trivial to define \cite{reinke2017average}. Goal-only rewards, defined as unit reward for reaching a goal and zero elsewhere, cause classic exploration techniques based on random-walk such as $\epsilon$-greedy and control input noise \cite{schulman2015trust}, or optimistic initialization to become highly inefficient. For example, Boltzmann exploration \cite{kaelbling1996reinforcement} requires training time exponential in the number of states \cite{osband2014generalization}. Such data requirement is unacceptable in real-world applications.
Most solutions to this problem rely on redesigning rewards to avoid dealing with the problem of exploration.
Reward shaping helps learning~\cite{ng1999policy}, and translating rewards to negative values triggers \textit{optimism in the face of uncertainty} \cite{kearns2002near,brafman2002r,jaksch2010near}. This approach suffers from two shortcomings: proper reward design is difficult and requires expert knowledge; improper reward design often degenerates to unexpected learned behaviour.

\begin{figure*}
\centering
    \includegraphics[width=.8\columnwidth]{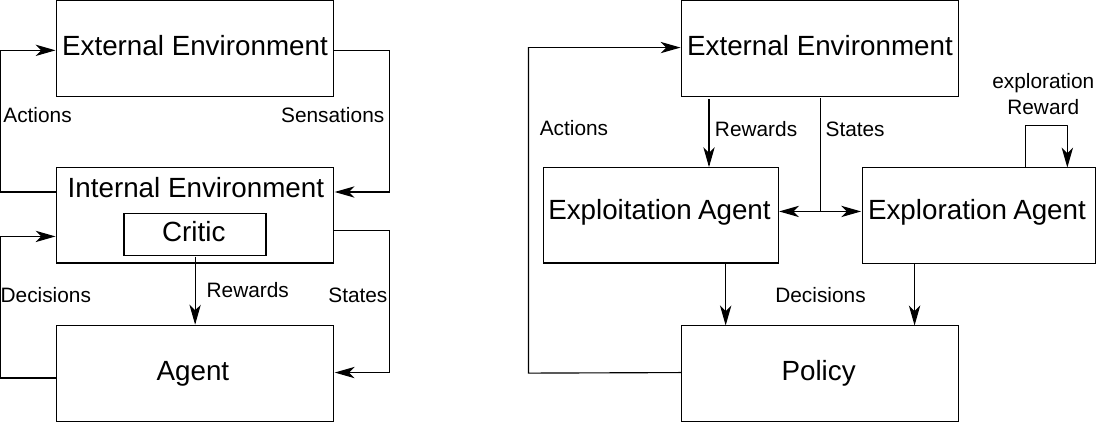}
	\caption{Left: classic intrinsic exploration setup as proposed in \cite{chentanez2005intrinsically}. Right: intrinsic exploration formulated as multi-objective RL\label{fig:intrinsicRLDiagram}}
\end{figure*}

Intrinsic motivation proposes a different approach to exploration by defining an additional guiding reward; see Figure~\ref{fig:intrinsicRLDiagram} (left). The exploration reward is typically added to the original reward, which makes rewards dense from the agent's perspective. This approach has had many successes \cite{bellemare2016unifying,fox2018dora} but suffers several limitations. For example, weighting between exploration and exploitation must be chosen before learning and remain fixed. Furthermore, in the model-free setting, state-action value functions are learned from non-stationary targets mixing exploration and exploitation, hence making learning less data-efficient.

To solve the problem of data-efficient exploration in goal-only reward settings, we propose to leverage advances in multi-objective RL \cite{Roijers2013}. We formulate exploration as one of the core objectives of RL by explicitly integrating it to the loss being optimized. Following the multi-objective RL framework, agents optimize for both exploration and exploitation as separate objectives. This decomposition can be seen as two different RL agents, as shown in Figure~\ref{fig:intrinsicRLDiagram} (right). Contrary to most intrinsic RL approaches, this formulation keeps the exploration-exploitation trade-off at a policy level, as in traditional RL. This allows for several advantages:
(i) Weighting between objectives can be adapted while learning, and strategies can be developed to change exploration online; (ii) Exploration can be stopped at any time at no extra cost, yielding purely exploratory behaviour immediately; (iii) Inspection of exploration status is possible, and experimenters can easily generate trajectories for exploration or exploitation only.

Our contributions are the following:
\begin{itemize}
    \item We propose a framework based on multi-objective RL for treating exploration as an explicit objective, making it core to the optimization problem.
    \item This framework is experimentally shown to perform better than classic additive exploration bonuses on several key exploration characteristics.
    \item Drawing inspiration from the fields of bandits and Bayesian optimization, we give strategies for taking advantage of and tuning the exploration-exploitation balance online. These strategies achieve a degree of control over agent exploration that was previously unattainable with classic additive intrinsic rewards.
    \item We present a data-efficient model-free RL method (EMU-Q) for continuous state-action goal-only MDPs based on the proposed framework, guiding exploration towards regions of higher value-function uncertainty.
    \item EMU-Q is experimentally shown to outperform classic exploration techniques and other methods with additive intrinsic rewards on a continuous control benchmark.
\end{itemize}

In the following, Section~\ref{sec:background} reviews background on Markov decision processes, intrinsic motivation RL, multi-objective RL and related work.
Section~\ref{sec:explicitBalanceExlp} defines a framework for explicit exploration-exploitation balance at a policy level, based on multi-objective RL.
Section~\ref{sec:controlKappa} presents advantages and strategies for controlling this balance during the agent learning process.
Section~\ref{sec:emuq} formulates EMU-Q, a model-free data-efficient RL method for continuous state-action goal-only MDPs, based on the proposed framework.
Section~\ref{sec:experiments} presents experiments that evaluate EMU-Q's exploration capabilities on classic RL problems and a simulated robotic manipulator. EMU-Q is further evaluated against other intrinsic RL methods on a continuous control benchmark.
We conclude with a summary in Section~\ref{sec:conclusion}.

\section{Preliminaries}
\label{sec:background}
This section reviews basics on Markov decision processes, intrinsic motivation RL, multi-objective RL and related work.
\subsection{Markov Decision Processes}
A Markov decision process (MDP) is defined by the tuple $<S,A,T,R,\gamma>$. $\mathcal{S}$ and $\mathcal{A}$ are spaces of states $s$ and actions $a$ respectively. The transition function $T: \mathcal{S} \times \mathcal{A} \times \mathcal{S} \rightarrow [0,1]$ encodes the probability to transition to state $s'$ when executing action $a$ in state $s$, i.e. $T(s,a,s')=p(s'|s,a)$. The reward distribution $R$ of support $\mathcal{S} \times \mathcal{A} \times \mathcal{S}$ defines the reward $r$ associated with transition $(s, a, s')$.
In the simplest case, goal-only rewards are deterministic and unit rewards are given for absorbing goal states, potential negative unit rewards are given for penalized absorbing states, and zero-reward is given elsewhere. $\gamma \in [0,1)$ is a discount factor.
Solving a MDP is equivalent to finding the optimal policy $\pi^*$ starting from $s_0$:
\begin{equation}
\label{eq:optimalPolicy}
	\pi^* = \arg\max_\pi \mathbb{E}_{T,R,\pi}[\sum_{i=0}^{\infty} \gamma^i r_i],
\end{equation}
with $a_{i} \sim \pi(s_i)$, $s_{i+1} \sim T(s_i, a_i, \cdot)$, and $r_i \sim R(s_i, a_i, s_{i+1})$.
Model-free RL learns an action-value function $Q$, which encodes the expected long-term discounted value of a state-action pair
\begin{equation}
\label{eq:qfunction}
Q(s,a) = \mathbb{E}_{T,R,\pi}[\sum_{i=0}^{\infty} \gamma^i r_i].
\end{equation}
Equation~\ref{eq:qfunction} can be rewritten recursively, also known as the Bellman equation
\begin{equation}
\label{eq:simplifiedBellman}
Q(s,a) = \mathbb{E}_R[R(s,a,s')] + \gamma \mathbb{E}_{s',a'|s,a}[Q(s',a')],
\end{equation}
$s' \sim p(s'|s,a)$, $a' \sim \pi(s')$, which is used to iteratively refine models of $Q$ based on transition data.

\subsection{Intrinsic RL}
While classic RL typically carries out exploration by adding randomness at a policy level (eg. random action, posterior sampling), intrinsic RL focuses on augmenting rewards with an exploration bonus. This approach was presented in \cite{chentanez2005intrinsically}, in which agents aim to maximize a total reward $r_{total}$ for transition $(s,a,r,s')$:
\begin{equation}
\label{eq:additiveReward}
	r_{total} = r + \xi r^e,
\end{equation}
where $r^e$ is the exploration bonus and $\xi$ a user-defined parameter weighting exploration.
The second term encourages agents to select state-action pairs for which they previously received high exploration bonuses. The definition of $r^e$ has been the focus of much recent theoretical and applied work; examples include model prediction error \cite{stadie2015incentivizing} or information gain \cite{little2013learning}.

While this formulation enables exploration in well behaved scenarios, it suffers from multiple limitations:
\begin{itemize}
    \item Exploration bonuses are designed to reflect the information gain at a given time of the learning process. They are initially high, and typically decrease after more transitions are experienced, making it a non-stationary target. Updating $Q$ with non-stationary targets results in higher data requirements, especially when environment rewards are stationary.
    \item The exploration bonus given for reaching new areas of the state-action space persists in the estimate of $Q$. As a consequence, agents tend to over-explore and may be stuck oscillating between neighbouring states.
    \item There is no dynamic control over the exploration-exploitation balance, as changing parameter $\xi$ only affects future total rewards. Furthermore, it would be desirable to control generating trajectories for pure exploration or pure exploitation, as these two quantities may conflict.
\end{itemize}
This work presents a framework for enhancing intrinsic exploration, which does not suffer from the previously stated limitations.

\subsection{Multi-Objective RL}
Multi-objective RL seeks to learn policies solving multiple competing objectives by learning how to solve for each objective individually \cite{Roijers2013}. In multi-objective RL, the reward function describes a vector of $n$ rewards instead of a scalar. The value function also becomes a vector $\bm{Q}$ defined as
\begin{equation}
    \bm{Q}(s,a) = \mathbb{E}_{T,R,\pi}[\sum_{i=0}^{\infty} \gamma^i \bm{r}_i],
\end{equation}
where $\bm{r}_i$ is the vector of rewards at step $i$ in which each coordinate corresponds to one objective. For simplicity, the overall objective is often expressed as the sum of all individual objectives; $\bm{Q}$ can be converted to a scalar state-action value function with a linear scalarization function: $Q_{\bm{\omega}}(s,a) = \bm{\omega}^T \bm{Q}(s,a)$, where $\bm{\omega}$ are weights governing the relative importance of each objective.

The advantage of the multi-objective RL formulation is to allow learning policies for all combinations of $\bm{\omega}$, even if the balance between each objective is not explicitly defined prior to learning. Moreover, if $\bm{\omega}$ is a function of time, policies for new values of $\bm{\omega}$ are available without additional computation. Conversely, with traditional RL methods, a pass through the whole dataset of transitions would be required.


\subsection{Related Work}
Enhancing exploration with additional rewards can be traced back to the work of \cite{storck1995reinforcement} and \cite{meuleau1999exploration}, in which information acquisition is dealt with in an active manner.
This type of exploration was later termed \emph{intrinsic motivation} and studied in \cite{chentanez2005intrinsically}. This field has recently received much attention, especially in the context of very sparse or \emph{goal-only} rewards \cite{reinke2017average,morere2018bayesian} where traditional reward functions give too little guidance to RL algorithms.

Extensive intrinsic motivation RL work has focused on domains with simple or discrete spaces, proposing various definitions for exploration bonuses.
Starting from reviewing intrinsic motivation in psychology, the work of \cite{oudeyer2008can} presents a definition based on information theory.
Maximizing predicted information gain from taking specific actions is the focus of \cite{little2013learning}, applied to learning in the absence of external reward feedback.
Using approximate value function variance as an exploration bonus was proposed in \cite{osband2016deep}.
In the context of model-based RL, exploration based on model learning progress \cite{lopes2012exploration}, and model prediction error \cite{stadie2015incentivizing,pmlr-v70-pathak17a} were proposed.
State visitation counts have been widely investigated, in which an additional model counting previous state-action pair occurrences guides agents towards less visited regions. Recent successes include \cite{bellemare2016unifying,fox2018dora}.
An attempt to generalizing counter-based exploration to continuous state spaces was made in \cite{nouri2009multi}, by using regression trees to achieve multi-resolution coverage of the state space.
Another pursuit for scaling visitation counters to large and continuous state spaces was made in \cite{bellemare2016unifying} by using density models.

Little work attempted to extend intrinsic exploration to continuous action spaces.
A policy gradient RL method was presented in \cite{houthooft2016vime}.
Generalization of visitation counters is proposed in \cite{fox2018dora}, and interpreted as exploration values.
Exploration values are also presented as an alternative to additive rewards in \cite{szita2008many}, where exploration balance at a policy level is mentioned.

Most of these methods typically suffer from high data requirements. One of the reasons for such requirements is that exploration is treated as an ad-hoc problem instead of being the focus of the optimization method.
More principled ways to deal with exploration can be found in other related fields.
In bandits, the balance between exploration and exploitation is central to the formulation \cite{kuleshov2014algorithms}.
For example with upper confidence bound \cite{auer2002finite}, actions are selected based on the balance between action values and a visitation term measuring the variance in the estimate of the action value.
In the bandits setting, the balance is defined at a policy level, and the exploration term is \emph{not} incorporated into action values like in intrinsic RL.

Similarly to bandits, Bayesian Optimization \cite{jones1998efficient} (BO) brings exploration at the core of its framework, extending the problem to continuous action spaces.
BO provides a data-efficient approach for finding the optimum of an unknown objective. Exploration is achieved by building a probabilistic model of the objective from samples, and exploiting its posterior variance information. An acquisition function such as UCB \cite{cox1992statistical} balances exploration and exploitation, and is at the core of the optimization problem.
BO was successfully applied to \emph{direct} policy search \cite{brochu2010tutorial,wilson2014using} by searching over the space of policy parameters, casting RL into a supervised learning problem.
Searching the space of policy parameters is however not data-efficient as recently acquired step information is not used to improve exploration. Furthermore, using BO as global search over policy parameters greatly restricts parameter dimensionality, hence typically imposes using few expressive and hand-crafted features.

In both bandits and BO formulations, exploration is brought to a policy level where it is a central goal of the optimization process.
In this work, we treat exploration and exploitation as two distinct objectives to be optimized. Multi-objective RL \cite{Roijers2013} provides tools which we utilize for defining these two distinct objectives, and balancing them at a policy level. Multi-objective RL allows for making exploration central to the optimization process.
While there exist Multi-objective RL methods to find several viable objective weightings such as finding Pareto fronts \cite{perny2010finding}, our work focuses on two well defined objectives whose weighting changes during learning. As such, we are mostly interested in the ability to change the relative importance of objectives without requiring training.

Modelling state-action values using a probabilistic model enables reasoning about the whole distribution instead of just its expectation, giving opportunities for better exploration strategies.
Bayesian Q-learning \cite{dearden1998bayesian} was first proposed to provide value function posterior information in the tabular case, then extended to more complicated domains by using Gaussian processes to model the state-action function \cite{engel2005reinforcement}. In this work, authors also discuss decomposition of returns into several terms separating intrinsic and extrinsic uncertainty, which could later be used for exploration.
Distribution over returns were proposed to design risk-sensitive algorithms \cite{morimura2010nonparametric}, and approximated to enhance RL stability in \cite{bellemare2017distributional}.
In recent work, Bayesian linear regression is combined to a deep network to provide a posterior on Q-values \cite{azizzadenesheli2018efficient}. Thomson sampling is then used for action selection, but can only guarantee local exploration. Indeed, if all action were experienced in a given state, the uncertainty of Q in this state is not sufficient to drive the agent towards unexplored regions.

To the best of our knowledge, there exists no model-free RL framework treating exploration as a core objective. We present such framework, building on theory from multi-objective RL, bandits and BO. We also present EMU-Q, a solution to exploration based on the proposed framework in fully continuous goal-only domains, relying on reducing the posterior variance of value functions.

This paper extends our earlier work \cite{morere2018bayesian}. It formalizes a new framework for treating exploration and exploitation as two objectives, provides strategies for online exploration control and new experimental results.

\section{Explicit Balance for Exploration and Exploitation}
\label{sec:explicitBalanceExlp}
Traditional RL aims at finding a policy maximizing the expected sum of future discounted rewards, as formulated in Equation~\ref{eq:optimalPolicy}. Exploration is then typically achieved by adding a perturbation to rewards or behaviour policies in an ad-hoc way.
We propose making the trade-off between exploration and exploitation explicit and at a policy level, by formulating exploration as a multi-objective RL problem.

\subsection{Framework Overview} 
Multi-objective RL extends the classic RL framework by allowing value functions or policies to be learned for individual objectives. Exploitation and exploration are two distinct objectives for RL agents, for which separate value functions $Q$ and $U$ (respectively) can be learned.
Policies then need to make use of information from two separate models for $Q$ and $U$. While exploitation value function $Q$ is learned from external rewards, exploration value function $U$ is modelled using exploration rewards.

Aiming to define policies which combine exploration and exploitation, we draw inspiration from Bayesian Optimization \cite{brochu2010tutorial}, which seeks to find the maximum of an expensive function using very few samples. It relies on an acquisition function to determine the most promising locations to sample next, based on model posterior mean and variance.
The Upper-Confidence Bounds (UCB) acquisition function \cite{cox1992statistical} is popular for its explicit balance between exploitation and exploration controlled by parameter $\kappa \in [0,\infty)$.
Adapting UCB to our framework leads to policies balancing $Q$ and $U$. Contrary to most intrinsic RL approaches, our formulation keeps the exploration-exploitation trade-off at a policy level, as in traditional RL.
This allows for adapting the exploration-exploitation balance during the learning process without sacrificing data-efficiency, as would be the case with a balance at a reward level.
Furthermore, policy level balance can be used to design methods to control the agent's learning process, e.g. stop exploration after a budget is reached, or encourage more exploration if the agent converged to a sub-optimal solution; see Section~\ref{sec:controlKappa}.
Lastly, generating trajectories resulting only from exploration or exploration grants experimenters insight over learning status.

\subsection{Exploration Values}
We propose to redefine the objective optimized by RL methods to incorporate both exploration and exploitation at its core. To do so, we consider the following expected \emph{balanced} return for policy $\pi$:
\begin{equation}
    \label{eq:balancedReturn}
    D^{\pi}(s,a) = \mathbb{E}_{T,R,R^e,\pi}[\sum_{i=0}^{\infty} \gamma^i (r_i + \kappa r^e_i)],
\end{equation}
where we introduced exploration rewards $r^e_i \sim R^e(s_i, a_i, s_{i+1})$ and parameter $\kappa \in [0,\infty]$ governing exploration-exploitation balance. Note that we recover Equation~\ref{eq:optimalPolicy} by setting $\kappa$ to $0$, hence disabling exploration.

Equation~\ref{eq:balancedReturn} can be further decomposed into
\begin{align}
\label{eq:balancedReturnSplit}
    D^{\pi}(s,a) &= \mathbb{E}_{T,R,\pi}[\sum_{i=0}^{\infty} \gamma^ir_i] + \kappa\mathbb{E}_{T,R^e,\pi}[\sum_{i=0}^{\infty} \gamma^ir^e_i]\\
            &= Q^\pi(s,a) + \kappa U^\pi(s,a),
\end{align}
where we have defined the exploration state-action value function $U$, akin to $Q$. Exploration behaviour is achieved by maximizing the expected discounted exploration return $U$. Note that, if $r^e$ depends on $Q$, then $U$ is a function of $Q$. For clarity, we omit this potential dependency in notations.

Bellman-type updates for both $Q$ and $U$ can be derived by unrolling the first term in both sums:

\begin{align}
    D^{\pi}(s,a) &= \mathbb{E}_R[r] + \mathbb{E}_{a',s'|s,a}[\mathbb{E}_{T,R,\pi}[\sum_{i=1}^{\infty} \gamma^ir_i]] + \kappa(\mathbb{E}_{R^e}[r^e] + \mathbb{E}_{a',s'|s,a}[\mathbb{E}_{T,R^e,\pi}[\sum_{i=1}^{\infty} \gamma^ir^e_i]])\\
                 &= \mathbb{E}_R[r] + \gamma\mathbb{E}_{a',s'|s,a}[Q^\pi(s',a')] + \kappa(\mathbb{E}_{R^e}[r^e] + \gamma\mathbb{E}_{a',s'|s,a}[U^\pi(s',a')]).
\end{align}
By identification we recover the update for $Q$ given by Equation~\ref{eq:simplifiedBellman} and the following update for $U$:
\begin{equation}
\label{eq:bellmanUpdateU}
U(s,a) = \mathbb{E}_{R^e}[r^e] + \gamma \mathbb{E}_{s',a'|s,a}[U(s',a')],
\end{equation}
which is similar to that of $Q$. Learning both $U$ and $Q$ can be seen as combining two agents to solve separate MPDs for goal reaching and exploration, as shown in Figure \ref{fig:intrinsicRLDiagram} (right). This formulation is general in that any reinforcement learning algorithm can be used to learn $Q$ and $U$, combined with any exploration bonus.
Both state-action value functions can be learned from transition data using existing RL algorithms.

\subsection{Exploration Rewards}
The presented formulation is independent from the choice of exploration rewards, hence many reward definitions from the intrinsic RL literature can directly be applied here.

Note that in the special case $R^e = 0$ for all states and actions, we recover exploration values from DORA \cite{fox2018dora}, and if state and action spaces are discrete, we recover visitation counters \cite{bellemare2016unifying}.

Another approach to define $R^e$ consists in considering the amount of exploration left at a given state. The exploration reward $r^{e}$ for a transition is then defined as the amount of exploration in the resulting state of the transition, to favour transitions that result in discovery. It can be computed by taking an expectation over all actions:
\begin{equation}
    \label{eq:explorationReward}
    R^e(s') = \mathbb{E}_{a'\sim \mathcal{U}(\mathcal{A})}[\sigma(s',a')].
\end{equation}
We defined a function $\sigma$ accounting for the uncertainty associated with a state-action pair. This formulation favours transitions that arrive at states of higher uncertainty.
An obvious choice for $\sigma$ is the variance of $Q$-values, to guide exploration towards parts of the state-action space where $Q$ values are uncertain. This formulation is discussed in Section~\ref{sec:emuq}.
Another choice for $\sigma$ is to use visitation count or its continuous equivalent. Compared to classic visitation counts, this formulation focuses on visitations of the resulting transition state $s'$ instead of on the state-action pair of origin $(s,a)$.

Exploration rewards are often constrained to negative values so that by combining an optimistic model for $U$ to negative rewards, \textit{optimism in the face of uncertainty} guarantees efficient exploration \cite{kearns2002near}. The resulting model creates a gradient of $U$ values; trajectories generated by following this gradient reach unexplored areas of the state-action space.
With continuous actions, Equation~\ref{eq:explorationReward} might not have closed form solution and the expectation can be estimated with approximate integration or sampling techniques. In domains with discrete actions however, the expectation is replaced by a sum over all possible actions.

\subsection{Action Selection} 
Goal-only rewards are often defined as deterministic, as they simply reflect goal and penalty states. Because our framework handles exploration in a deterministic way, we simply focus on deterministic policies.
Although state-action values $Q$ are still non-stationary (because $\pi$ is), they are learned from a stationary objective $r$. This makes learning policies for exploitation easier.

Following the definition in Equation~\ref{eq:balancedReturn}, actions are selected to maximize the expected balanced return $D^\pi$ at a given state $s$:
\begin{equation}
\label{eq:boPolicy}
\pi(s) = \arg\max_a D^\pi(s,a) = \arg\max_a Q^\pi(s,a) + \kappa U^\pi(s,a).
\end{equation}
Notice the similarity between the policy given in Equation~\ref{eq:boPolicy} and UCB acquisition functions from the Bayesian optimization and bandits literature. No additional exploration term is needed, as this policy explicitly balances exploration and exploitation with parameter $\kappa$.
This parameter can be tuned at any time to generate trajectories for pure exploration or exploitation, which can be useful to assess agent learning status. Furthermore, strategies can be devised to control $\kappa$ manually or automatically during the learning process. We propose a few strategies in Section~\ref{sec:controlKappa}.

The policy from Equation~\ref{eq:boPolicy} can further be decomposed into present and future terms:
\begin{equation}
\pi(s) = \arg\max_a \underbrace{\mathbb{E}_R[r] + \kappa\mathbb{E}_{R^e}[r^e]}_{myopic} + \gamma\mathbb{E}_{s'|s,a}[\underbrace{\mathbb{E}_{a'\sim\pi(s')}[Q^\pi(s',a') +  \kappa U^\pi(s',a')]}_{future}],
\end{equation}
where the term denoted \emph{future} is effectively $D^\pi(s')$. This decomposition highlights the link between this framework and other active learning methods; by setting $\gamma$ to $0$, only the \emph{myopic} term remains, and we recover the traditional UCB acquisition function from bandits or Bayesian optimization.
This decomposition can be seen as an extension of these techniques to a non-myopic setting. Indeed, future discounted exploration and exploitation are also considered within the action selection process.
Drawing this connection opens up new avenues for leveraging exploration techniques from the bandits literature.

The method presented in this section for explicitly balancing exploration and exploitation at a policy level is concisely summed up in Algorithm \ref{alg:explicitExplorationExploitation}. The method is general enough so that it allows learning both $Q$ and $U$ with any RL algorithm, and does not make assumptions on the choice of exploration reward used. Section~\ref{sec:emuq} presents a practical method implementing this framework, while the next section presents advantages and strategies for controlling exploration balance during the agent learning process.

\begin{algorithm}[t]
\caption{Explicit Exploration-Exploitation}
\label{alg:explicitExplorationExploitation}
\begin{algorithmic}[1]
\STATE \textbf{Input:} parameter $\kappa$.
\STATE \textbf{Output:} Policy $\pi$.
\FOR{episode $l=1,2, ..$}
\FOR{step $h=1, 2, ..$}
\STATE $\pi(s) = \arg\max_{a\in\mathcal{A}} Q(s,a) + \kappa U(s,a)$
\STATE Execute $a=\pi(s)$, observe $s'$ and $r$, and store $s, a, r, s'$ in $D$.
\STATE Generate $r^e$ with Equation~\ref{eq:explorationReward} for example.
\STATE Update $Q$ with Bellman eq. and $r$.
\STATE Update $U$ with Bellman eq. and $r^e$.
\ENDFOR
\ENDFOR
\end{algorithmic}
\end{algorithm}
\section{Preliminary Experiments on Classic RL Problems}
\label{sec:controlKappa}
In this section, a series of preliminary experiments on goal-only classic RL domains is presented to highlight the advantages of exploration values over additive rewards. Strategies for taking advantage of variable exploration rates are then provided.

The comparisons make use of goal-only version of simple and fully discrete domains. We compare all methods using strictly the same learning algorithm and reward bonuses. Learning algorithms are tabular implementations of Q-Learning with learning rate fixed to $0.1$. Reward bonuses are computed from a table of state-action visitation counts, where experiencing a state-action pair for the first time grants $0$ reward and revisiting yields $-1$ reward.
We denote by \textit{additive reward} a learning algorithm where reward bonuses are used as in classic intrinsic RL (Equation~\ref{eq:additiveReward}), and by \textit{exploration values} reward bonuses used as in the proposed framework (Equation~\ref{eq:bellmanUpdateU} and action selection defined by Equation~\ref{eq:boPolicy}). A Q-learning agent with no reward bonuses and $\epsilon$-greedy exploration is displayed as a baseline.

{\it Problem 1:} The Cliff Walking domain \cite{sutton1998reinforcement} is adapted to the goal-only setting: negative unit rewards are given for falling off the cliff (triggering agent teleportation to starting state), and positive unit rewards for reaching the terminal goal state. Transitions allow the agent to move in four cardinal directions, where a random direction is chosen with low probability $0.01$.

{\it Problem 2:} The traditional Taxi domain \cite{dietterich2000hierarchical} is also adapted to the goal-only setting. This domain features a $5\times5$ grid-word with walls and four special locations. In each episode, the agent starts randomly and two of the special locations are denoted as passenger and destination. The goal is for the agent to move to the passenger's location, pick-up the passenger, drive it to the destination, and drop it off. A unit reward is given for dropping-off the passenger to the destination (ending the episode), and $-0.1$ rewards are given for actions \textit{pick-up} and \textit{drop-off} in wrong locations.

\subsection{Analyzing the Advantages of an Explicit Exploration-Exploitation Balance}
We first present simple pathological cases in which using exploration values provides advantages over additive rewards for exploration, on the Cliff Walking domain.

\begin{figure*}[t]
\centering
\begin{minipage}[c]{\textwidth}
	\subfloat[][]{
   \includegraphics[width=0.48\columnwidth]{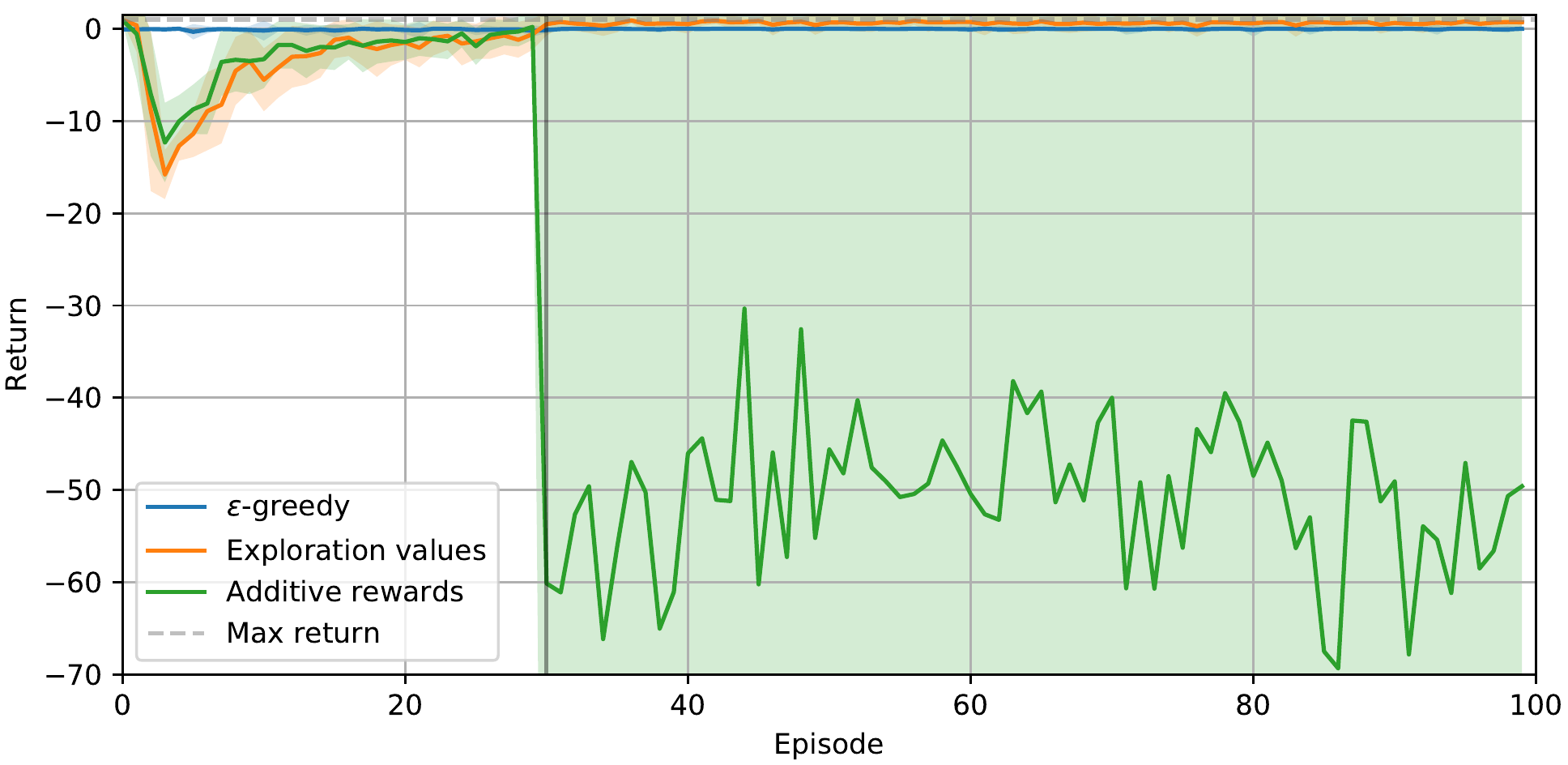}
   \label{fig:explorationStopping}
 }
 \subfloat[][]{
   \includegraphics[width=0.48\columnwidth]{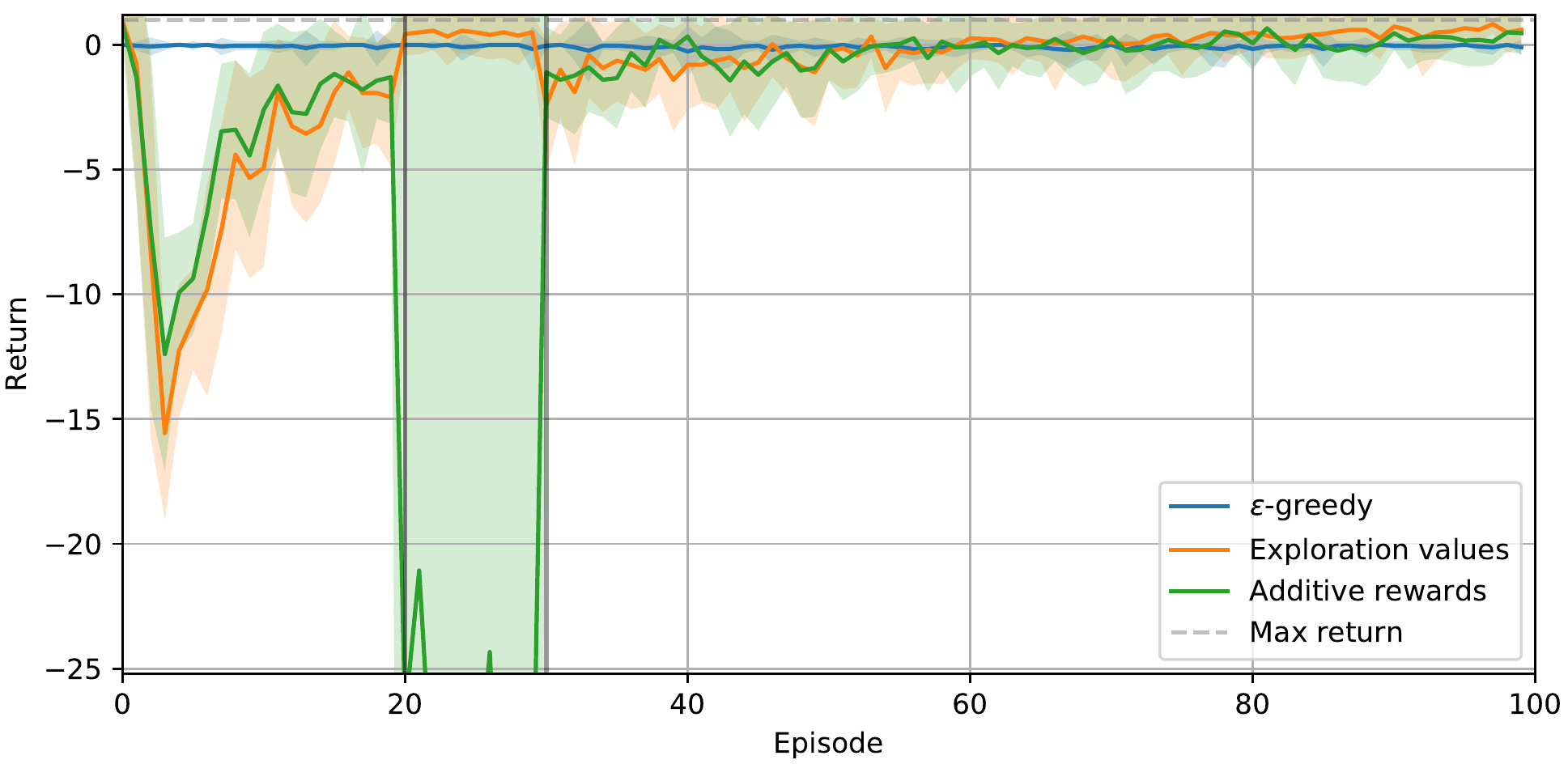}
   \label{fig:explorationStopContinue}
 }
\end{minipage}\\
\begin{minipage}[c]{\textwidth}
 \subfloat[][]{
   \includegraphics[width=0.48\columnwidth]{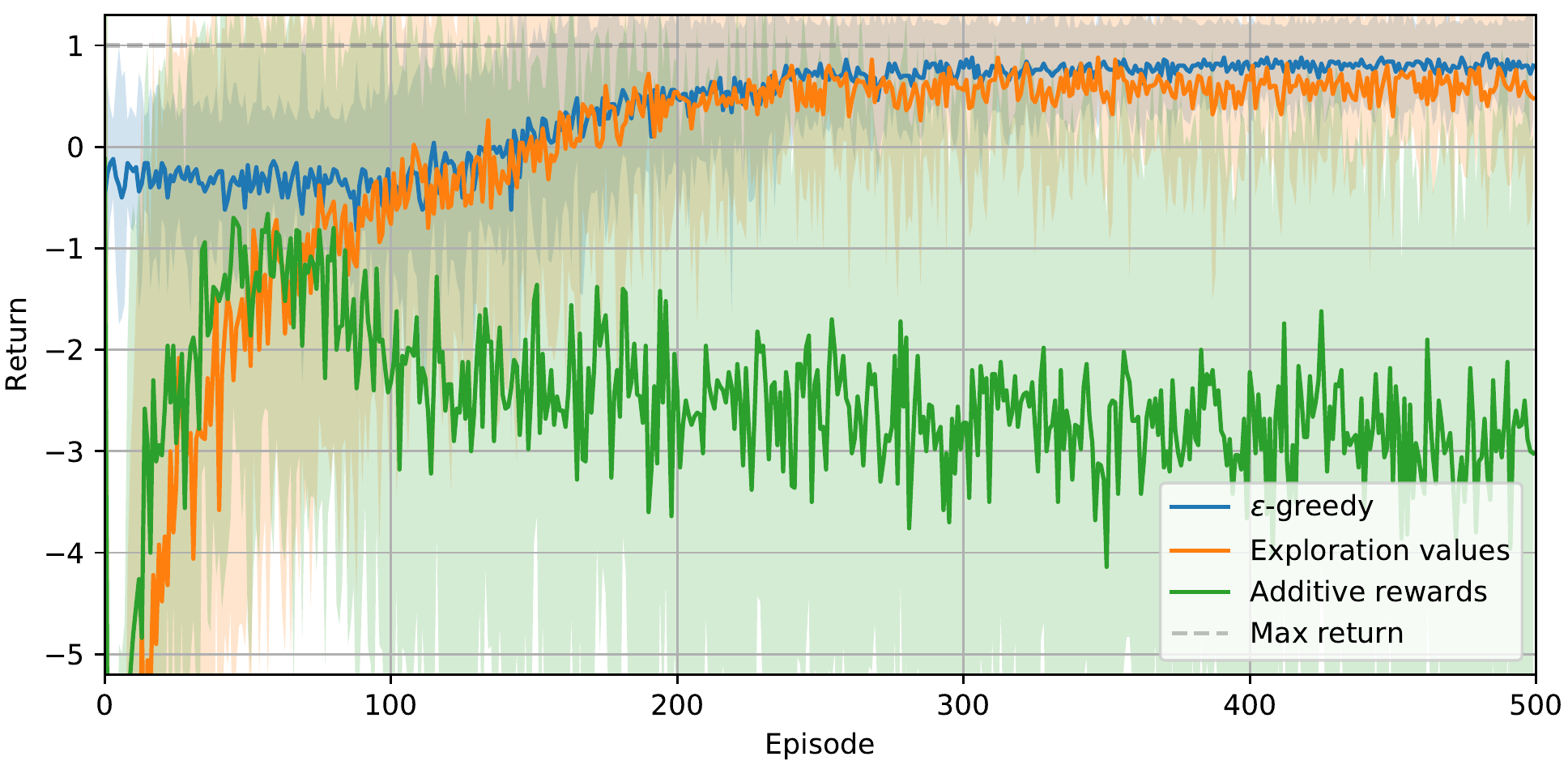}
   \label{fig:explorationStochasticT}
 }
 \subfloat[][]{
   \includegraphics[width=0.48\columnwidth]{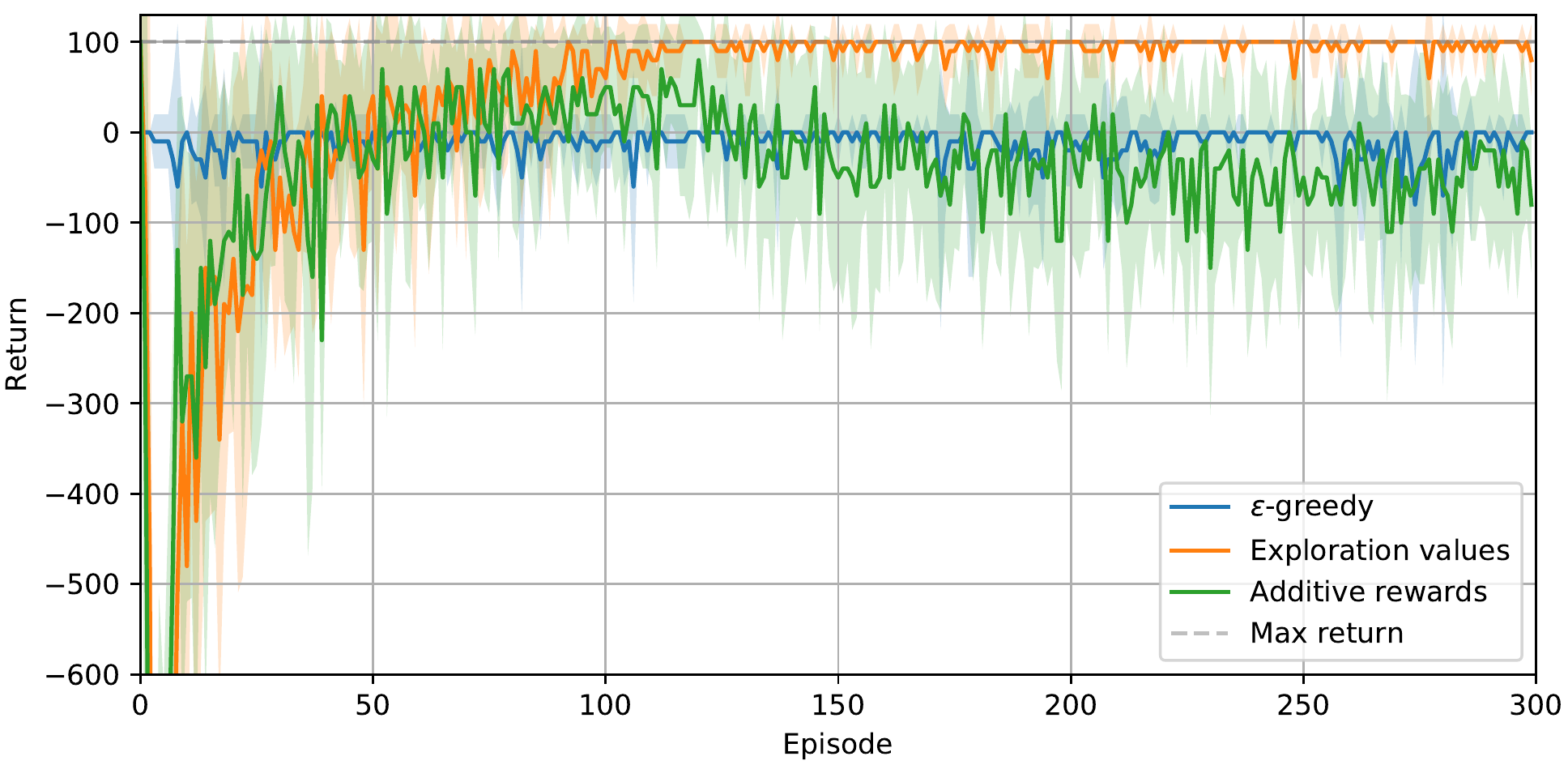}
   \label{fig:explorationRewardAmpl}
 }
\end{minipage}
\caption{Mean return and variance of $100$ runs of the Cliff Walking domain with goal-only rewards. (a) Stopping exploration after 30 episodes. (b) Stopping exploration after 20 episodes and continuing exploration after another 10 episodes. (c) Stochastic transitions with $0.1$ probability of random action. (d) Rewards are scaled by $100$, and exploration parameters are also scaled to keep an equal magnitude between exploration and exploitation terms.\label{fig:cliffWalkingComparison}}
\end{figure*}

The first two experiments show that exploration values allow for direct control over exploration such stopping and continuing exploration.
Stopping exploration after a budget is reached is simulated by setting exploration parameters (eg. $\kappa$) to $0$ after $30$ episodes and stopping model learning. While exploration values maintain high performance after exploration stops, returns achieved with additive rewards dramatically drop and yield a degenerative policy. When exploration is enabled once again, the two methods continue improving at a similar rate; see Figures~\ref{fig:explorationStopping} and \ref{fig:explorationStopContinue}.
Note that when exploration is disabled, there is a jump in returns with exploration values, as performance for pure exploitation is evaluated. However, it is never possible to be sure a policy is generated from pure exploitation when using additive rewards, as parts of bonus exploration rewards are encoded within learned Q-values.

The third experiment demonstrates that stochastic transitions with higher probability of random action ($p=0.1$) lead to increased return variance and poor performance with additive rewards, while exploration values only seem mildly affected. As shown in Figure~\ref{fig:explorationStochasticT}, even $\epsilon$-greedy appears to solve the task, suggesting stochastic transitions provide additional random exploration.
It is unclear why the additive rewards method is affected negatively. 

\begin{figure*}[t]
\centering
\begin{minipage}[c]{\textwidth}
	\subfloat[][]{
   \includegraphics[width=0.48\columnwidth]{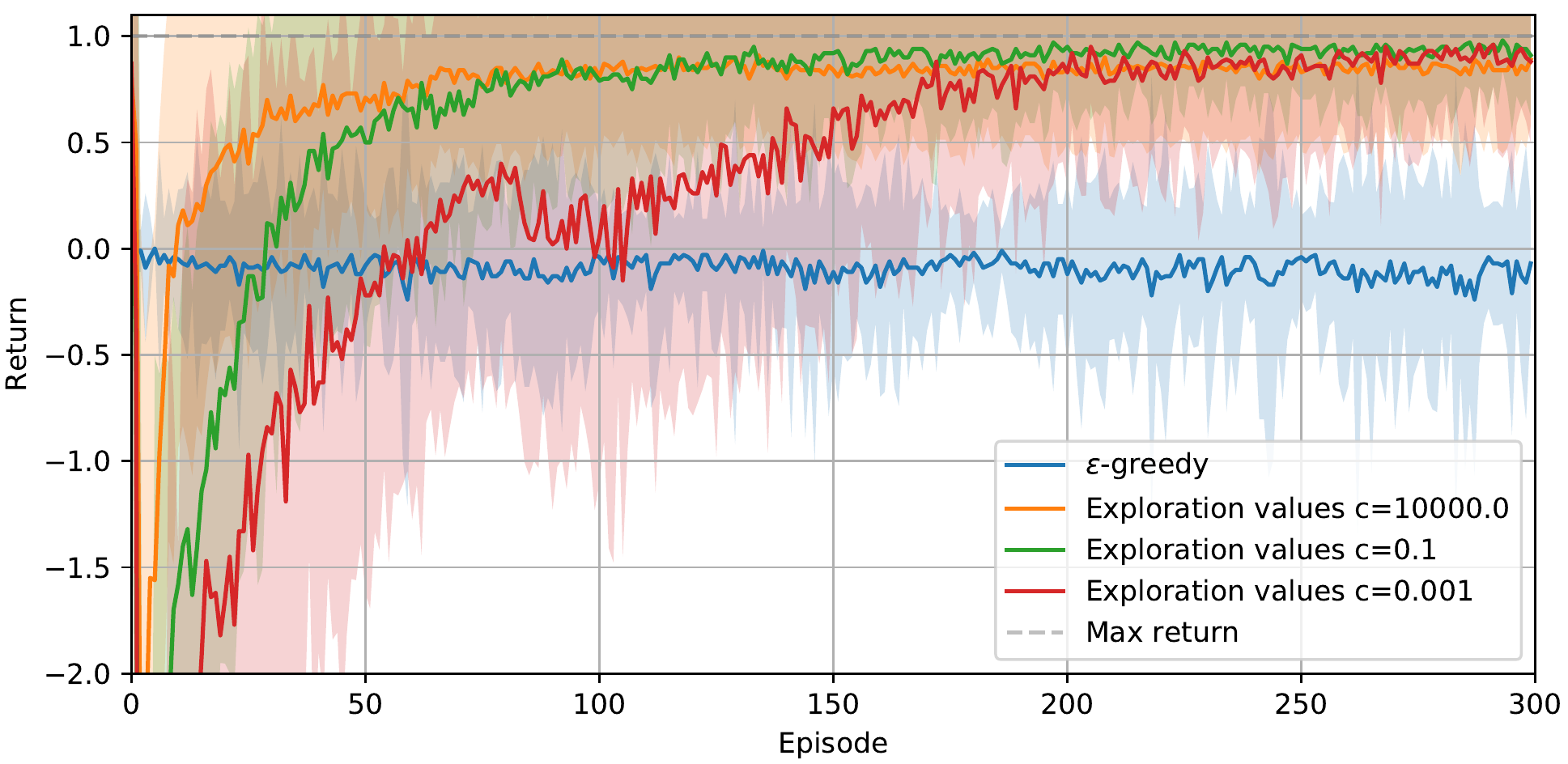}   
 }
 \subfloat[][]{
   \includegraphics[width=0.48\columnwidth]{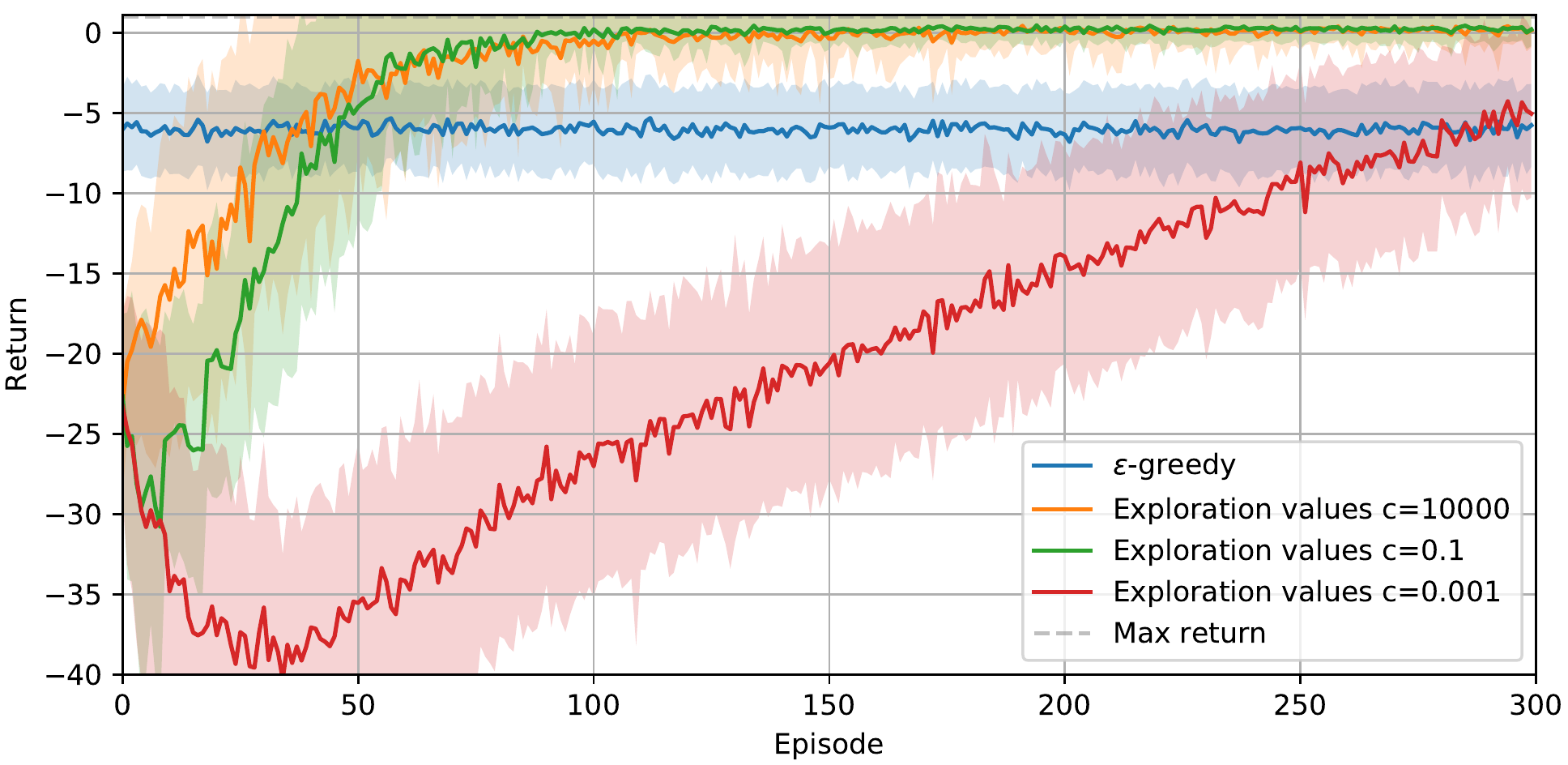}
 }
\end{minipage}
\caption{Decreasing exploration parameter over time to control exploration level on sparse Cliff Walking (a) and sparse Taxi (b) domains. Results show return mean and variance of a 100 runs.\label{fig:decreasedExpl}}
\end{figure*}

Lastly, the fourth experiments shows environment reward magnitude is paramount to achieving good performance with additive rewards; see Figure~\ref{fig:explorationRewardAmpl}. Even though exploration parameters balancing environment and exploration bonus rewards are scaled to maintain equal amplitude between the two terms, additive rewards suffer from degraded performance. This is due to two reward quantities being incorporated into a single model for Q, which also needs to be initialized optimistically with respect to both quantities. When the two types of rewards have different amplitude, this causes a problem. Exploration values do not suffer from this drawback as separate models are learned based on these two quantities, hence resulting in unchanged performance.

\subsection{Automatic Control of Exploration-Exploitation Balance}
We now present strategies for automatically controlling the exploration-exploitation balance during the learning process. The following experiments also make use of the Taxi domain.

Exploration parameter $\kappa$ is decreased over time according to the following schedule $\kappa (t) = \frac{1}{1+ct}$, where $c$ governs decay rate. Higher values of $c$ result in reduced exploration after only a few episodes, whereas lower values translate to almost constant exploration. Results displayed in Figure~\ref{fig:decreasedExpl} indicate that decreasing exploration leads to fast convergence to returns relatively close to maximum return, as shown when setting $c=10^5$. However choosing a reasonable value $c=0.1$ first results in lower performance, but enables finding a policy with higher returns later. Such behaviour is more visible with very small values such as $c=10^{-3}$ which corresponds to almost constant $\kappa$.

\begin{table}
	\centering
	\begin{tabular}{ l | c c c}
		\textbf{Method} & Times target reached & Episodes to target & Performance after target\\
	\hline
	$\epsilon$-greedy & 0/10 & -- & --\\
	Exploration values & 9/10 & \textbf{111.33} & 	\textbf{0.08}(4.50)\\
	Additive rewards & 9/10 & 242.11 & -33.26(67.41)\\
\end{tabular}
\caption{Stopping exploration after a target test return of $0.1$ is reached on $5$ consecutive episodes in the sparse Taxi domain. Results averaged over 100 runs.\label{tab:explTargetReturn}}
\end{table}

\begin{figure*}[t]
\centering
\begin{minipage}[c]{\textwidth}
	\subfloat[][]{
   \includegraphics[width=0.48\columnwidth]{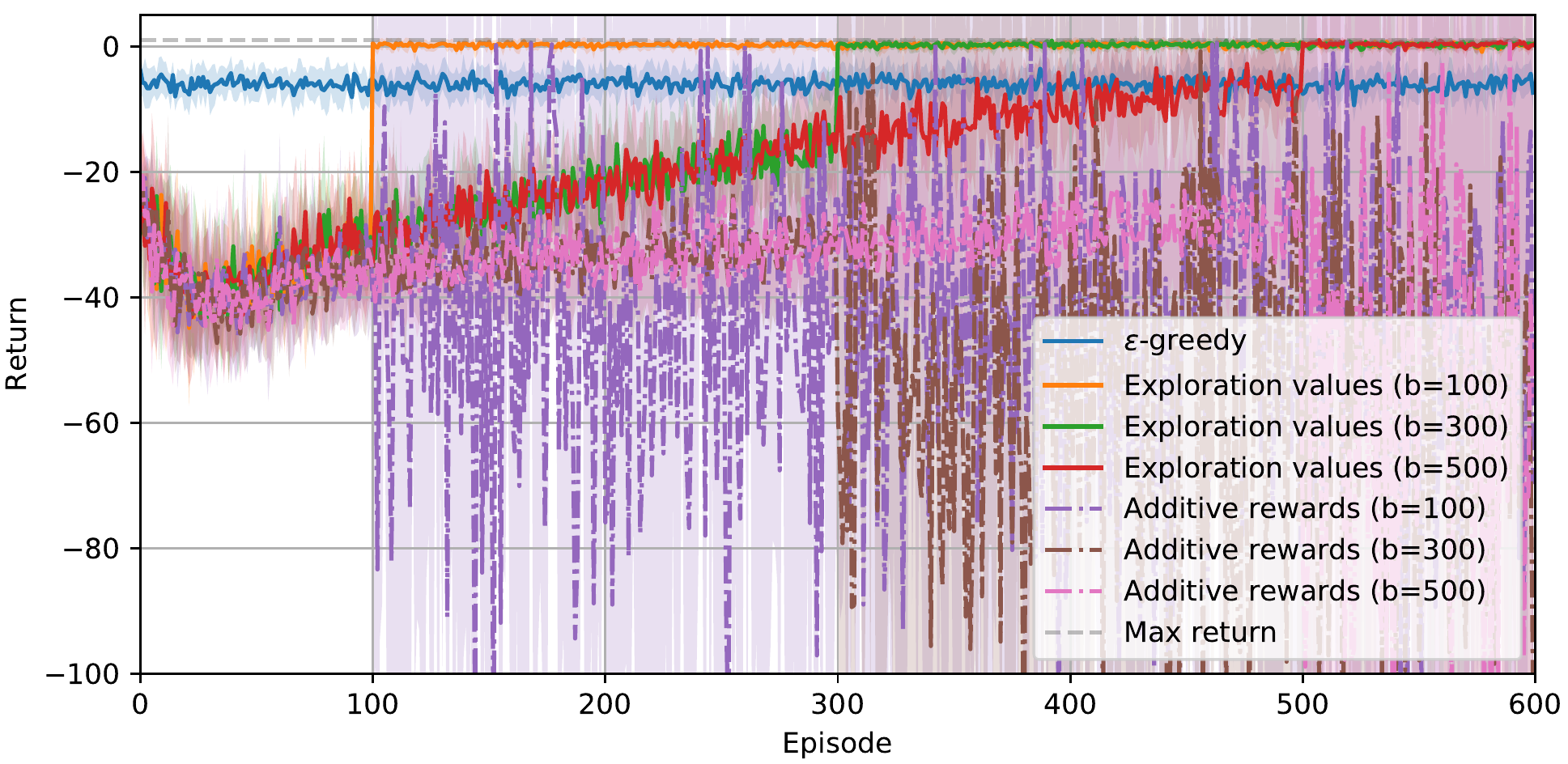}
   \label{fig:explStopBudget}
 }
 \subfloat[][]{
   \includegraphics[width=0.48\columnwidth]{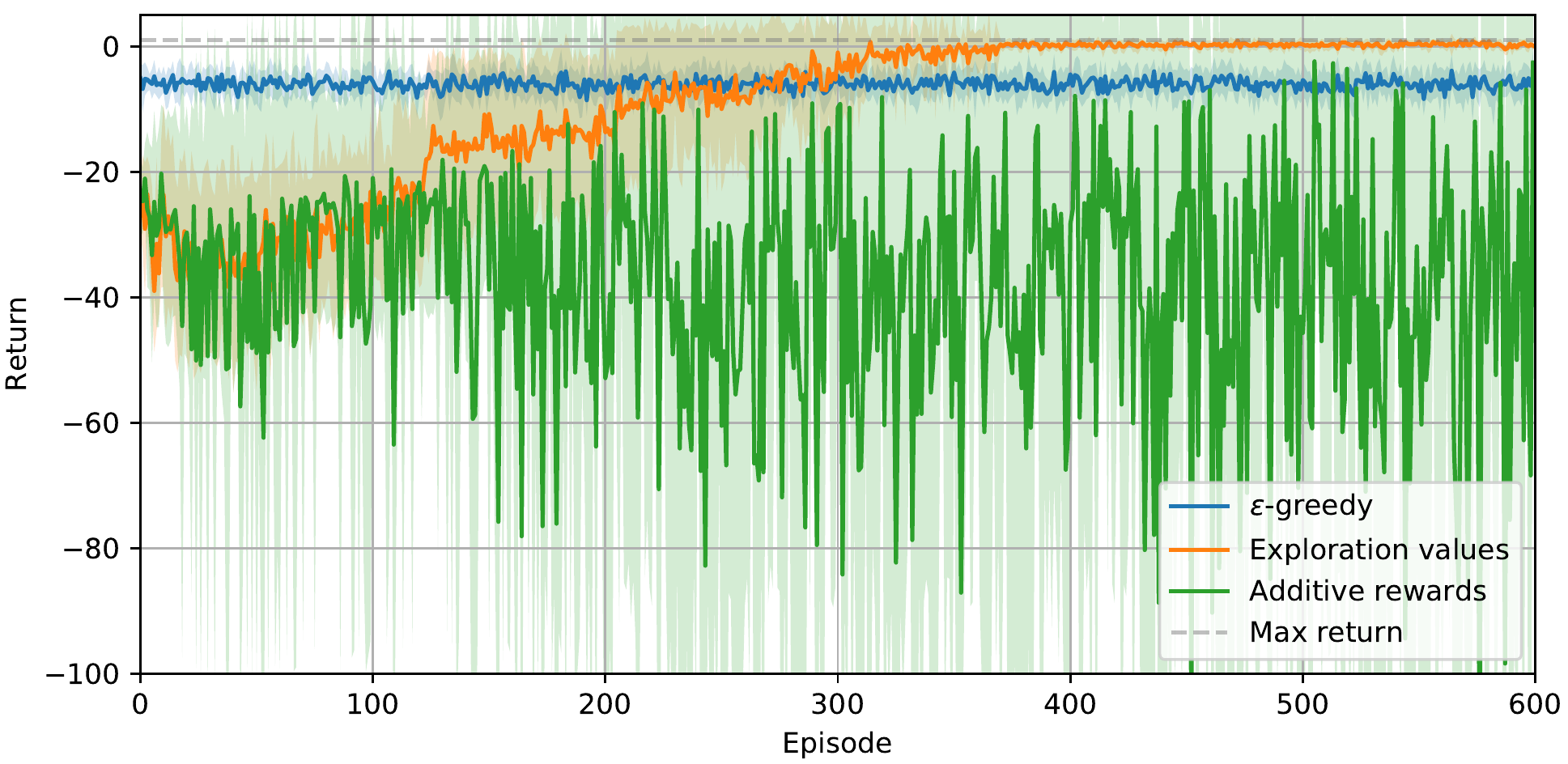}
   \label{fig:explStopReturn}
 }
\end{minipage}
\caption{Taxi domain with goal-only rewards. (a) Exploration stops after a fixed budget of episodes $b \in \{100, 300, 500\}$ is exhausted. (b) Exploration stops after a test target return of $0.1$ is reached on $5$ consecutive runs. Results show return mean and variance of a 100 runs.}
\end{figure*}

We now show how direct control over exploration parameter $\kappa$ can be taken advantage of to stop learning automatically once a predefined target is met. On the taxi domain, exploration is first stopped after an exploration budget is exhausted. Results comparing additive rewards to exploration values for different budgets of $100$, $300$ and $500$ episodes are given in Figure~\ref{fig:explStopBudget}. These clearly show that when stopping exploration after the budget is reached, exploration value agents can generate purely exploiting trajectories achieving near optimal return whereas additive reward agents fail to converge on an acceptable policy.

Lastly, we investigate stopping exploration automatically once a target return is reached. After each learning episode, $5$ test episodes with pure exploitation are run to score the current policy. If all $5$ test episodes score returns above $0.1$, the target return is reached and exploration stops. Results for this experiment are shown in Figure~\ref{fig:explStopReturn} and Table~\ref{tab:explTargetReturn}. Compared to additive rewards, exploration values display better performance after target is reach as well as faster target reaching.


Exploration values were experimentally shown exploration advantages over additive reward on simple RL domains. The next section presents an algorithm built on the proposed framework which extends to fully continuous state and action spaces and is applicable to more advanced problems.

\section{EMU-Q: Exploration by Minimizing Uncertainty of $Q$ Values}
\label{sec:emuq}
Following the framework defined in Section \ref{sec:explicitBalanceExlp}, we propose learning exploration values with a specific reward driving trajectories towards areas of the state-action space where the agent's uncertainty of $Q$ values is high.
\subsection{Reward Definition}
Modelling $Q$-values with a probabilistic model gives access to variance information representing model uncertainty in expected discounted returns. Because the probabilistic model is learned from \emph{expected} discounted returns, discounted return variance is not considered. Hence the model variance only reflects epistemic uncertainty, which can be used to drive exploration.
This formulation was explored in EMU-Q \cite{morere2018bayesian}, extending Equation~\ref{eq:explorationReward} as follows:
\begin{equation}
    \label{eq:explorationRewardEMUQ}
    R^e(s') = \mathbb{E}_{a'\sim \mathcal{U}(\mathcal{A})}[\mathbb{V}[Q(s',a')]] - \mathbb{V}_{max},
\end{equation}
where $\mathbb{V}_{max}$ is the maximum possible variance of $Q$, guaranteeing always negative rewards. In practice, $\mathbb{V}_{max}$ depends on the model used to learn $Q$ and its hyper-parameters, and can often be computed analytically.
In Equation~\ref{eq:explorationRewardEMUQ}, the variance operator $\mathbb{V}$ computes the \emph{epistemic} uncertainty of $Q$ values, that is it assesses how confident the model is that it can predict $Q$ values correctly. Note that the MDP stochasticity emerging from transitions, rewards and policy is absorbed by the expectation operator in Equation~\ref{eq:qfunction}, and so no assumptions are required on the MDP components in this reward definition.

\subsection{Bayesian Linear Regression for Q-Learning}
\label{sec:bayesianQLearning}
We now seek to obtain a model-free RL algorithm able to explore with few environment interactions, and providing a full predictive distribution on state-action values to fit the exploration reward definition given by Equation~\ref{eq:explorationRewardEMUQ}.
Kernel methods such as \emph{Gaussian Process TD} (GPTD) \cite{engel2005reinforcement} and \emph{Least-Squares TD} (LSTD) \cite{lagoudakis2003least} are among the most data-efficient model-free techniques. While the former suffers from prohibitive computation requirements, the latter offers an appealing trade-off between data-efficiency and complexity.
We now derive a Bayesian RL algorithm that combines the strengths of both kernel methods and LSTD.

The distribution of long-term discounted \emph{exploitation} returns $G$ can be defined recursively as:
\begin{equation}
	\label{eq:returnDistrib}
	G(s,a) = R(s,a,s') + \gamma G(s',a'),
\end{equation}
which is an equality in the distributions of the two sides of the equation. Note that so far, no assumption are made on the nature of the distribution of returns.
Let us decompose the discounted return $G$ into its mean $Q$ and a random zero-mean residual $q$ so that $Q(s,a) = \mathbb{E}[G(s,a)]$. Substituting and rearranging Equation~\ref{eq:returnDistrib} yields
\begin{equation}
	\underbrace{R(s,a,s') + \gamma Q(s',a')}_{t} = Q(s,a) + \underbrace{q(s,a) - \gamma q(s',a')}_{\epsilon}.
\end{equation}
The only \emph{extrinsic} uncertainty left in this equation are the reward distribution $R$ and residuals $q$. Assuming rewards are disturbed by zero-mean Gaussian noise implies the difference of residuals $\epsilon$ is Gaussian with zero-mean and precision $\beta$. By modelling $Q$ as a linear function of a feature map $\bm{\phi}_{s,a}$ so that $Q(s,a) = \bm{w}^T \bm{\phi}_{s,a}$, estimation of state-action values becomes a linear regression problem of target $\bm{t}$ and weights $\bm{w}$. The likelihood function takes the form
\begin{equation}
	p(\bm{t}|\bm{x},\bm{w},\beta) = \prod_{i=1}^N \mathcal{N}(t_i|r_i+\gamma\bm{w}^T\bm{\phi}_{s'_i,a'_i}, \beta^{-1}),
\end{equation}
where independent transitions are denoted as $x_i=(s_i,a_i,r_i,s'_i,a'_i)$. We now treat the weights as random variables with zero-mean Gaussian prior $p(\bm{w}) = \mathcal{N}(\bm{w}|\bm{0},\alpha^{-1}\bm{I})$. The weight posterior distribution is
\begin{align}
p(\bm{w}|\bm{t}) &= \mathcal{N}(\bm{w}|\bm{m}_Q,\bm{S})\\
	\bm{m}_Q &= \beta\bm{S}\bm{\Phi}_{\bm{s},\bm{a}}^T(\bm{r} + \gamma \bm{Q}')\\
	\bm{S} &= (\alpha\bm{I}+\beta\bm{\Phi}_{\bm{s},\bm{a}}^T\bm{\Phi}_{\bm{s},\bm{a}})^{-1},
\end{align}
where $\bm{\Phi}_{\bm{s},\bm{a}} = \{\bm{\phi}_{s_i,a_i}\}^N_{i=1}$, $\bm{Q}' = \{Q(s'_i,a'_i)\}^N_{i=1}$, and $\bm{r} = \{r_i\}^N_{i=1}$.
The predictive distribution is also Gaussian, yielding
\begin{align}
Q(s,a) = \mathbb{E}[p(t|\bm{x},\bm{t}, \alpha, \beta)] &= \bm{\phi}_{s,a}^T\bm{m}_Q,\\
\text{and }\mathbb{V}[p(t|\bm{x},\bm{t}, \alpha, \beta)] &= \beta^{-1} \bm{\phi}_{s,a}^T\bm{S}\bm{\phi}_{s,a}.
\end{align}
The predictive variance $\mathbb{V}[p(t|\bm{x},\bm{t}, \alpha, \beta)]$ encodes the \emph{intrinsic} uncertainty in $Q(s,a)$, due to the subjective understanding of the MDP's model; it is used to compute $r^{e}$ in Equation~\ref{eq:explorationRewardEMUQ}.

The derivation for $U$ is similar, replacing $r$ with $r^{e}$ and $t$ with $t^{e} = R^e(s,a,s') + \gamma U(s',a')$.
Note that because $\bm{S}$ does not depend on rewards, it can be shared by both models. Hence, with $\bm{U}' = \{U(s'_i,a'_i)\}^N_{i=1}$,
\begin{equation}
U(s,a) = \bm{\phi}_{s,a}^T\bm{m}_U, \text{ with }
\bm{m}_U = \beta\bm{S}\bm{\Phi}_{\bm{s},\bm{a}}^T(\bm{r}^{e} + \gamma \bm{U}').
\end{equation}

This model gracefully adapts to iterative updates at each step, by substituting the current prior with the previous posterior. Furthermore, the Sherman-Morrison equality is used to compute rank-1 updates of matrix $\bm{S}$ with each new data point $\phi_{s,a}$:
\begin{equation}
\bm{S}_{t+1} = \bm{S}_{t} - \beta \frac{(\bm{S}_{t} \bm{\phi}_{s,a})(\bm{\phi}_{s,a}^T \bm{S}_{t})}{1 + \beta \bm{\phi}_{s,a}^T \bm{S}_{t} \bm{\phi}_{s,a}}
\end{equation}
This update only requires a matrix-to-vector multiplication and saves the cost of inverting a matrix at every step. Hence the complexity cost is reduced from $O(M^3)$ to $O(M^2)$ in the number of features $M$. An optimized implementation of EMU-Q is given in algorithm \ref{alg:finalAlgo}.

\begin{algorithm}[t]
\caption{EMU-Q}
\label{alg:finalAlgo}
\begin{algorithmic}[1]
\STATE \textbf{Input:} initial state $s$, parameters $\alpha, \beta, \kappa$.
\STATE \textbf{Output:} Policy $\pi$ paramatrized by $\bm{m}_Q$ and $\bm{m}_U$.
\STATE Initialize $\bm{S} = \alpha^{-1}\bm{I}$, $\bm{t}_Q = \bm{t}_U = \bm{m}_Q = \bm{m}_U = \bm{0}$
\FOR{episode $l=1,2, ..$}
\FOR{step $h=1, 2, ..$}
\STATE $\pi(s) = \arg\max_a \bm{\phi}_{s,a} \bm{m}_Q + \kappa \bm{\phi}_{s,a} \bm{m}_U$
\STATE Execute $a=\pi(s)$, observe $s'$ and $r$, and store $\bm{\phi}_{s,a}, r, s'$ in $D$.
\STATE Generate $r^{e}$ from Equation~\ref{eq:explorationRewardEMUQ} with $s'$.
\STATE $\bm{S} = \bm{S} - \beta \frac{(\bm{S} \bm{\phi}_{s,a})(\bm{\phi}_{s,a}^T \bm{S})}{1 + \beta \bm{\phi}_{s,a}^T \bm{S} \bm{\phi}_{s,a}}$
\STATE $\bm{t}_Q = \bm{t}_Q + \beta \bm{\phi}_{s,a}^T (r + \gamma \bm{\phi}_{s,a} \bm{m}_Q)$
\STATE $\bm{t}_U = \bm{t}_U + \beta \bm{\phi}_{s,a}^T (r^{e} + \gamma \bm{\phi}_{s,a} \bm{m}_U)$.

\STATE $\bm{m}_Q = \bm{S}\bm{t}_Q$, $\bm{m}_U = \bm{S}\bm{t}_U$
\ENDFOR

\STATE From $D$, draw $\bm{\Phi}_{\bm{s},\bm{a}}, \bm{r}, \bm{s}'$, and compute $\bm{\Phi}_{\bm{s}',\pi(\bm{s}')}$.
\STATE Update $\bm{m}_Q = \beta \bm{S} \bm{\Phi}_{\bm{s},\bm{a}}^T (\bm{r} + \gamma \bm{\Phi}_{\bm{s}',\pi(\bm{s}')} \bm{m}_Q)$ until change in $\bm{m}_Q < \epsilon$.

\STATE Compute $\bm{r}^{e}$ with Equation~\ref{eq:explorationRewardEMUQ} and $\bm{s}'$.
\STATE Update $\bm{m}_U = \beta \bm{S} \bm{\Phi}_{\bm{s},\bm{a}}^T (\bm{r}^{e} + \gamma \bm{\Phi}_{\bm{s}',\pi(\bm{s}')} \bm{m}_U)$ until change in $\bm{m}_U < \epsilon$.

\ENDFOR
\end{algorithmic}
\end{algorithm}
End of episode updates for $\bm{m}_Q$ and $\bm{m}_U$ (line 15 onward) are analogous to policy iteration, and although not mandatory, greatly improve convergence speed. Note that because $r^{e}$ is a non-stationary target, recomputing it after each episode with the updated posterior on $Q$ provides the model on $U$ with more accurate targets, thereby improving learning speed.

\subsection{Kernel Approximation Features for RL}
We presented a simple method to learn $Q$ and $U$ as linear functions of states-actions features. While powerful when using a good feature map, linear models typically require experimenters to define meaningful features on a problem specific basis.
In this section, we introduce random Fourier features (RFF) \cite{rahimi2008random}, a kernel approximation technique which allows linear models to enjoy the expressivity of kernel methods. It should be noted that these features are different from Fourier basis \cite{konidaris2011value} (detailed in supplementary material), which do not approximate kernel functions. Although RFF were recently used to learn policy parametrizations \cite{rajeswaran2017towards}, to the best of our knowledge, this is the first time RFF are applied to the value function approximation problem in RL.

For any shift invariant kernel, which can be written as $k(\tau)$ with $\tau = \bm{x} - \bm{x}'$, a representation based on the Fourier transform can be computed with Bochner's theorem \cite{gihman1974th}.

\noindent
{\bf Theorem 1} {\it (Bochner's Theorem) Any shift invariant kernel $k(\tau)$, $\tau \in \mathbb{R}^D$, with a positive finite measure $d\mu(\bm{\omega})$ can be represented in terms of its Fourier transform as
	\begin{equation}
		k(\tau) = \int_{\mathbb{R}^D} e^{-i\bm{\omega}\tau} d\mu(\bm{\omega}).
	\end{equation}
} \hfill\BlackBox

\noindent
Assuming measure $\mu$ has a density $p(\bm{\omega})$, $p$ is the spectral density of $k$ and we have
\begin{equation}
k(\tau) = \int_{\mathbb{R}^D} e^{-i\tau\bm{\omega}}p(\bm{\omega})d\bm{\omega}
		\approx \frac{1}{M} \sum_{j=1}^{M} e^{-i\tau\bm{\omega}_j}
		= \langle \bm{\phi}(\bm{x}), \bm{\phi}(\bm{x}') \rangle,
\end{equation}
where $p$ is the spectral density of $k$, $\bm{\phi}(\bm{x})$ is an approximate feature map, and $M$ the number of spectral samples from $p$. In practice, the feature map approximating $k(\bm{x},\bm{x}')$ is
\begin{equation}
\label{eq:RFF}
\bm{\phi}(\bm{x}) = \frac{1}{\sqrt{M}} [\cos(\bm{x}^T\bm{\omega}_1), ..., \cos(\bm{x}^T\bm{\omega}_M),
										\sin(\bm{x}^T\bm{\omega}_1), ..., \sin(\bm{x}^T\bm{\omega}_M)],
\end{equation}
where the imaginary part was set to zero, as required for real kernels.
In the case of the RBF kernel defined as
$k(\bm{x},\bm{x}') = \exp(-\frac{1}{2\sigma^2}||\bm{x}-\bm{x}'||_2^2)$,
the kernel spectral density is Gaussian $p=\mathcal{N}(0, 2\sigma^{-2}I)$.
Feature maps can be computed by drawing $M/2 \times d$ samples from $p$ one time only, and computing Equation~\ref{eq:RFF} on new inputs $\bm{x}$ using these samples. Resulting features are not domain specific and require no feature engineering. Users only need to choose a kernel that represents adequate distance measures in the state-action space, and can benefit from numerous kernels already provided by the literature.
Using these features in conjunction with Bayesian linear regression provides an efficient method to approximate a Gaussian process.

As the number of features increases, kernel approximation error decreases \cite{sutherland2015error}; approximating popular shift-invariant kernels to within $\epsilon$ can be achieved with only $M= O(d\epsilon^{-2}\log\frac{1}{\epsilon^2})$ features. Additionally, sampling frequencies according to a quasi-random sampling scheme (used in our experiments) reduces kernel approximation error compared to classic Monte-Carlo sampling with the same number of features \cite{yang2014quasi}.

EMU-Q with RFF combines the ease-of-use and expressivity of kernel methods brought by RFF with the convergence properties and speed of linear models.

\subsubsection{Comparison of Random Fourier Features and Fourier Basis Features}
\begin{figure*}
    \centering
    \begin{minipage}[c]{.32\textwidth}
    	\subfloat[][]{
       \includegraphics[width=\columnwidth]{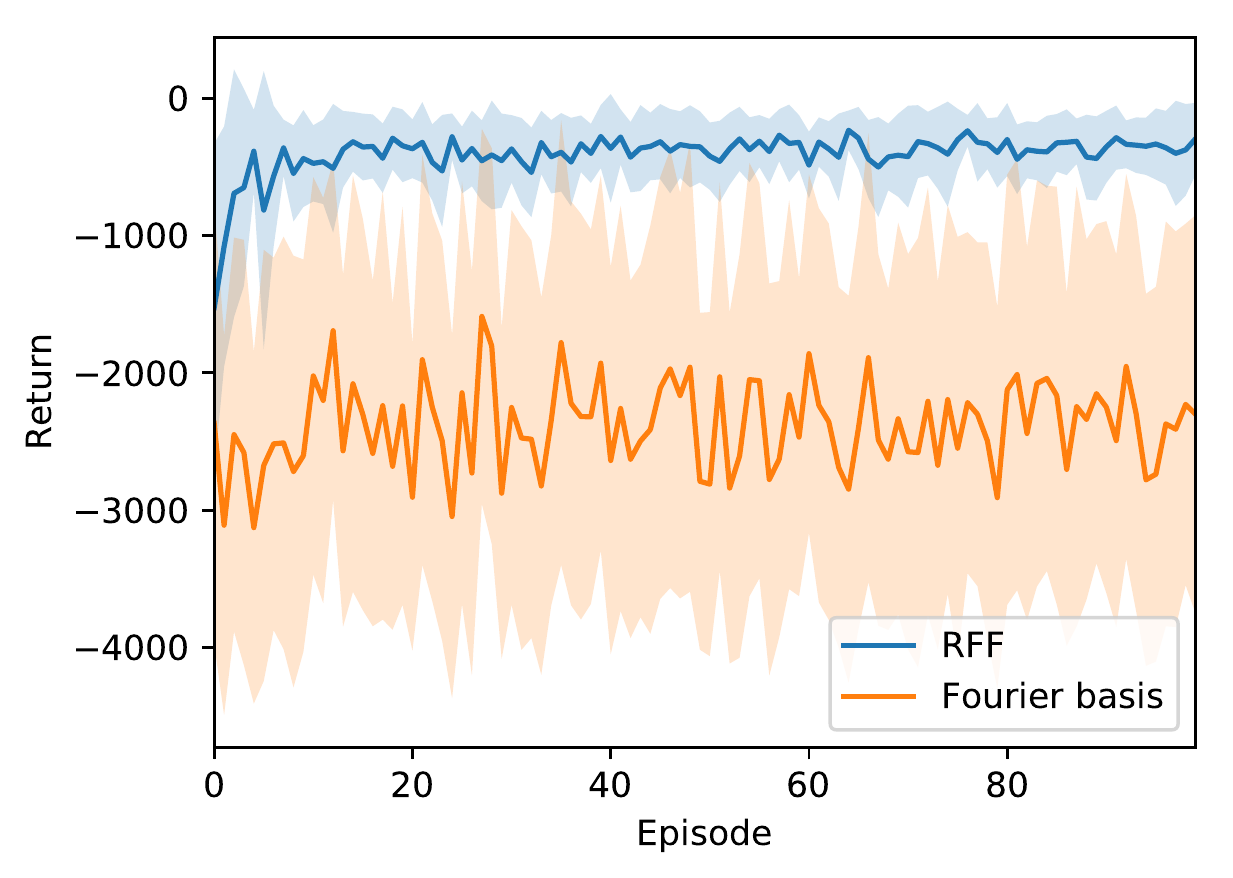}
     }
    \end{minipage}
    \begin{minipage}[c]{.32\textwidth}
     \subfloat[][]{
       \includegraphics[width=\columnwidth]{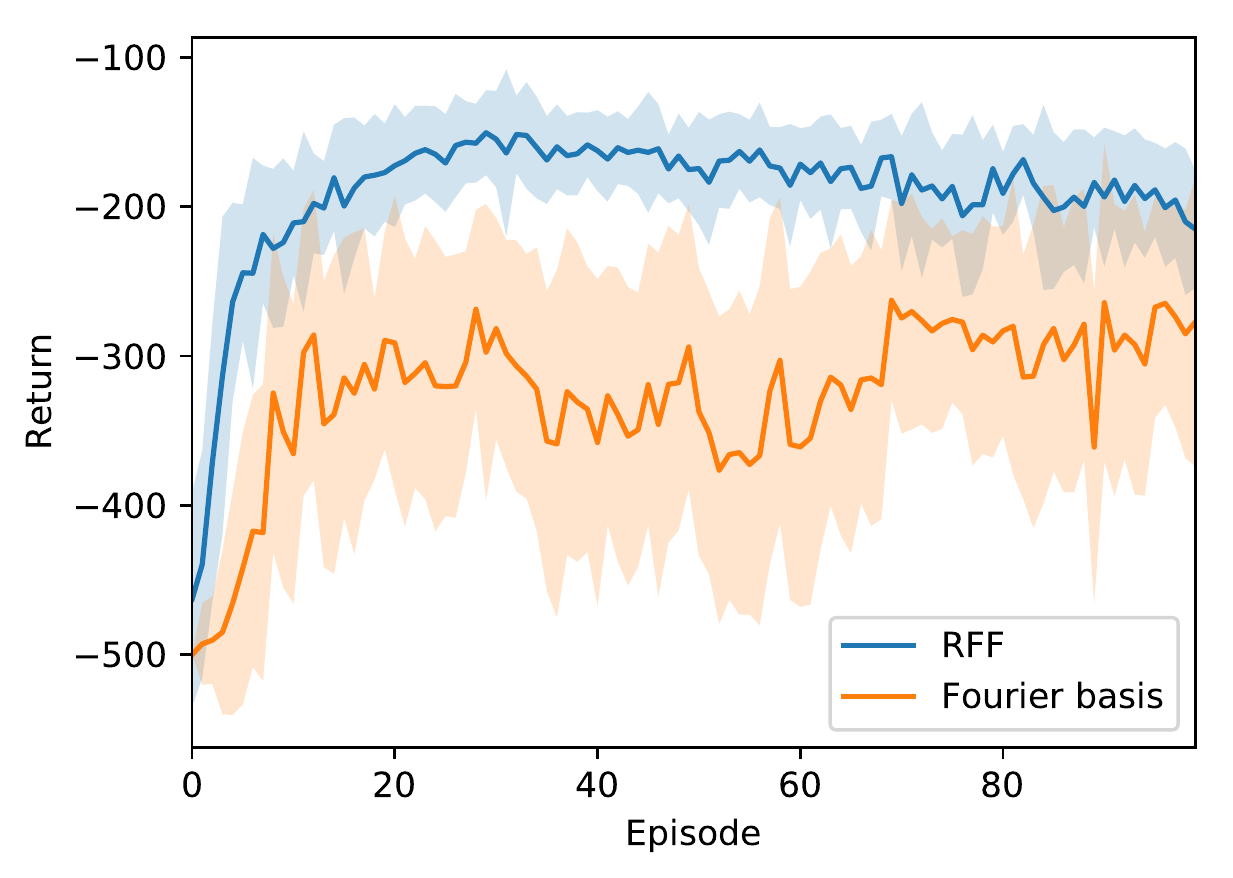}
     }
    \end{minipage}
    \begin{minipage}[c]{.32\textwidth}
     \subfloat[][]{
       \includegraphics[width=\columnwidth]{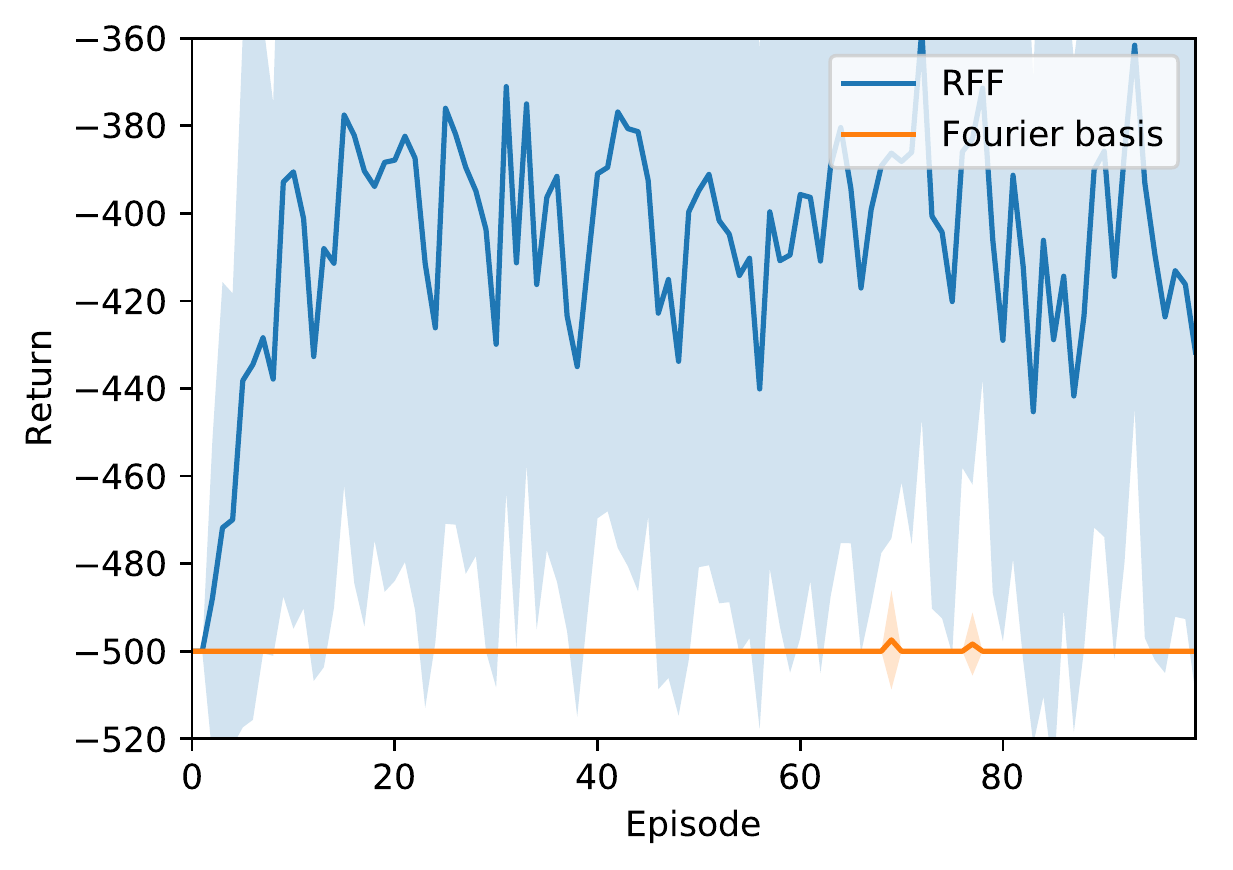}
     }
    \end{minipage}
    \caption{Return mean and standard deviation for Q-learning with random Fourier features (RFF) or Fourier basis features on SinglePendulum (a), MountainCar (b), and DoublePendulum (c) domain with classic rewards. Results are computed using classic Q-learning with $\epsilon$-greedy policy, and averaged over 20 runs for each method.\label{fig:comparisonRFF_FBF}}
\end{figure*}

For completeness, a comparison between RFF and the better known Fourier Basis Features \cite{konidaris2011value} is provided on classic RL domains using Q-learning. A short overview on Fourier Basis Features is given in Appendix~\ref{app:fourier_basis_features}.

Three relatively simple environments were considered: SinglePendulum, MountainCar and DoublePendulum (details on these environments are given in Section~\ref{sec:experiments}). The same Q-learning algorithm was used for both methods, with equal parameters.
As little as $300$ random Fourier features are sufficient in these domains, while the order of Fourier basis was set to $5$ for SinglePendulum and MountainCar and to $3$ for DoublePendulum. The higher state and action space dimensions of DoublePendulum make using Fourier basis features prohibitively expensive, as the number of generated features increases exponentially with space dimensions. For example, in DoublePendulum, Fourier basis features of order $3$ leads to more than $2000$ features.

Results displayed in Figure~\ref{fig:comparisonRFF_FBF} show RFF outperforms Fourier basis both in terms of learning speed and asymptotic performance, while using a lower number of features.
In DoublePendulum, the number of Fourier basis features seems insufficient to solve the problem, even though it is an order of magnitude higher than that of RFF.

\section{Experiments}
\label{sec:experiments}
EMU-Q's exploration performance is qualitatively and quantitatively evaluated on a toy chain MDP example, $7$ widely-used continuous control domains and a robotic manipulator problem. Experiments aim at measuring exploration capabilities in domains with goal-only rewards. Unless specified otherwise, domains feature one absorbing goal state with positive unit reward, and potential penalizing absorbing states of reward of $-1$. All other rewards are zero, resulting in very sparse reward functions, and rendering guidance from reward gradient information inapplicable.

\subsection{Synthetic Chain Domain}
\begin{figure*}
\centering
\begin{minipage}[c]{.35\textwidth}
 \subfloat[][]{
    \vbox to 8.8em{
    \vfil
    \includegraphics[width=\columnwidth]{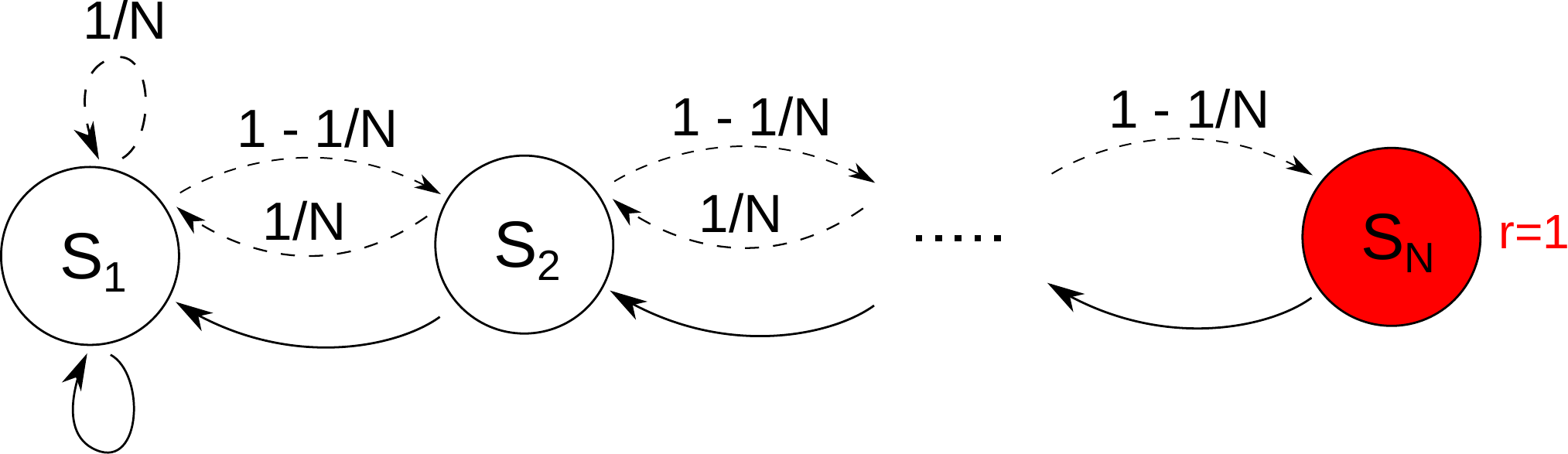}
    \label{fig:hardChain}
    \vfil}}
\end{minipage}
\begin{minipage}[c]{.55\textwidth}
 \subfloat[][]{
   \includegraphics[width=.5\columnwidth]{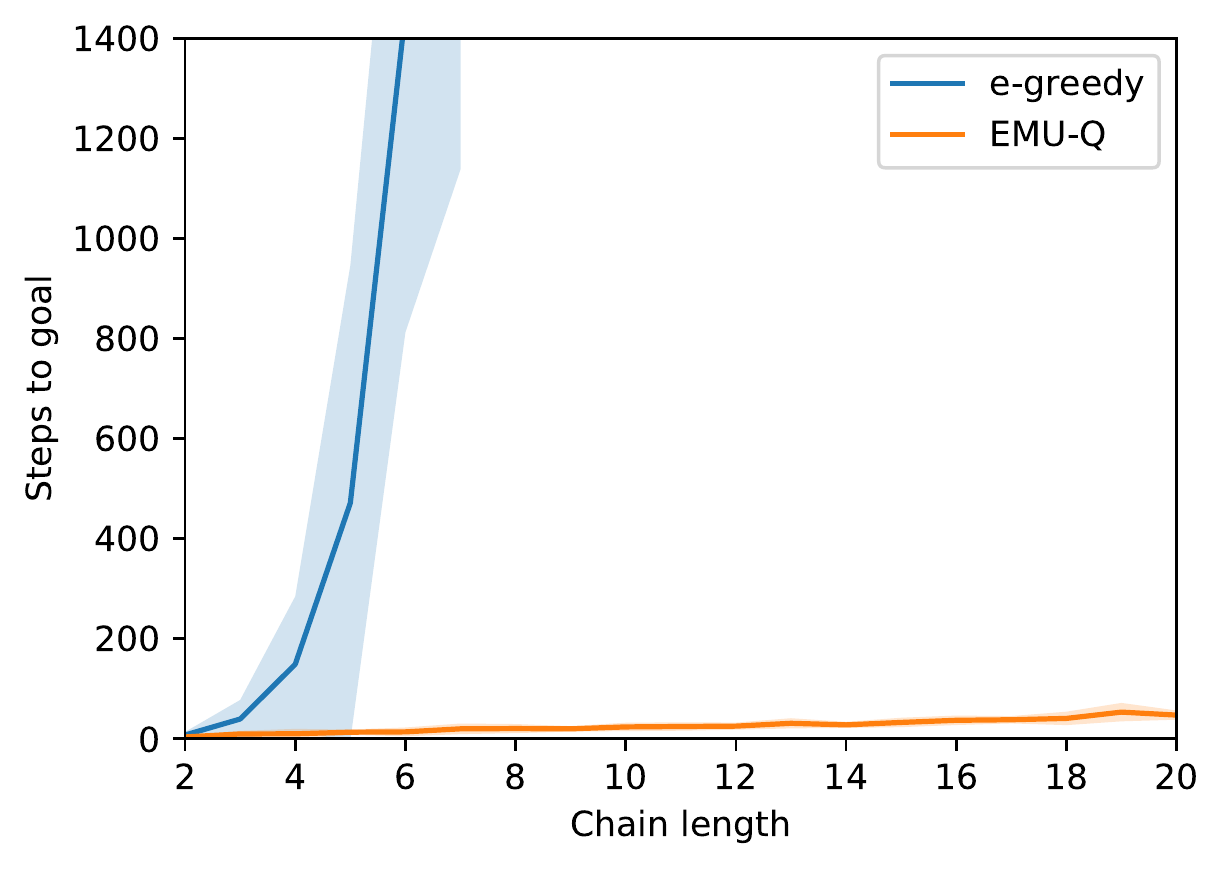}
   \label{fig:resultIncreasedChain}
 }
 \subfloat[][]{
   \includegraphics[width=.5\columnwidth]{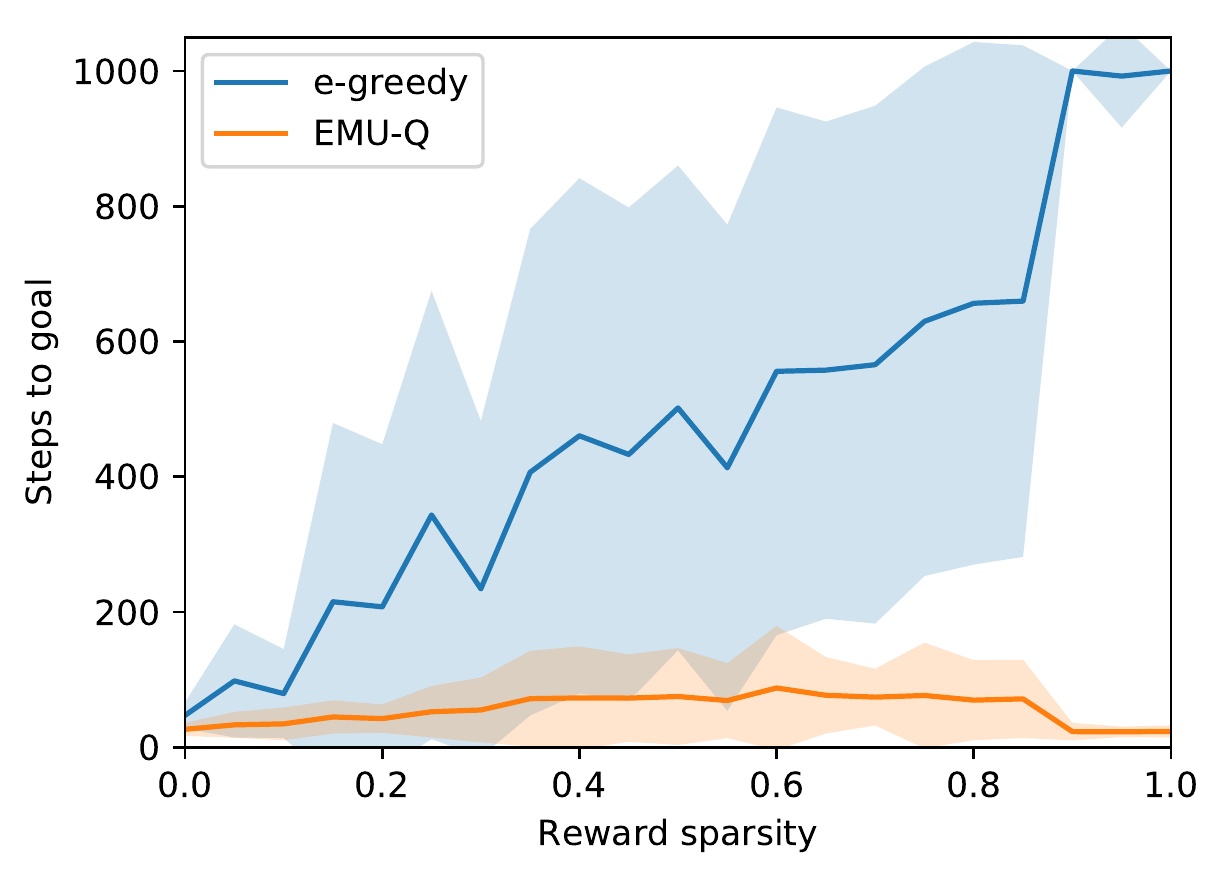}
   \label{fig:rewardSparsityChain}
 }
\end{minipage}
\caption{(a) Chain domain described in \cite{osband2014generalization}. (b) Steps to goal (mean and standard deviation) in chain domain, for increasing chain lengths, averaged over 30 runs. (c) Steps to goal in semi-sparse 10-state chain, as a function of reward sparsity, with maximum of 1000 steps (averaged over 100 runs).}
\end{figure*}

We investigate EMU-Q's exploration capabilities on a classic domain known to be hard to explore. It is composed of a chain of $N$ states and two actions, displayed in Figure~\ref{fig:hardChain}. Action right (dashed) has probability $1 - 1/N$ to move right and probability $1/N$ to move left. Action left (solid) is deterministic.

\subsubsection{Goal-only Rewards}
We first consider the case of goal-only rewards, where goal state $S_N$ yields unit reward and all other transitions result in nil reward.
Classic exploration such as $\epsilon$-greedy was shown to have exponential regret with the number of states in this domain \cite{osband2014generalization}. Achieving better performance on this domain is therefore essential to any advanced exploration technique.
We compare EMU-Q to $\epsilon$-greedy exploration for increasing chain lengths, in terms of number of steps before goal-state $S_N$ is found.
Results in Figure~\ref{fig:resultIncreasedChain} illustrate the exponential regret of $\epsilon$-greedy while EMU-Q achieves much lower exploration time, scaling linearly with chain length.

\subsubsection{Semi-Sparse Rewards}
We now investigate the impact of reward structure by decreasing the chain domain's reward sparsity. In this experiment \emph{only}, agents are given additional $-1$ rewards with probability $1-p$ for every non-goal state, effectively guiding them towards the goal state (goal-only rewards are recovered for $p=0$). The average number of steps before the goal is reached as a function of $p$ is compared for $\epsilon$-greedy and EMU-Q in Figure~\ref{fig:rewardSparsityChain}. Results show that $\epsilon$-greedy performs very poorly for high $p$, but improves as guiding reward density increases. Conversely, EMU-Q seems unaffected by reward density and performs equally well for all values of $p$. When $p=0$, agents receive $-1$ reward in every non-goal state, and $\epsilon$-greedy performs similarly to EMU-Q.

\subsection{Classic Control}
EMU-Q is further evaluated on more challenging RL domains. These feature fully continuous state and action spaces, and are adapted to the goal-only reward setting. In these standard control problems \cite{gym}, classic exploration methods are unable to reach goal states.

\begin{figure*}
\centering
	\subfloat[][]{
   \includegraphics[width=.23\columnwidth]{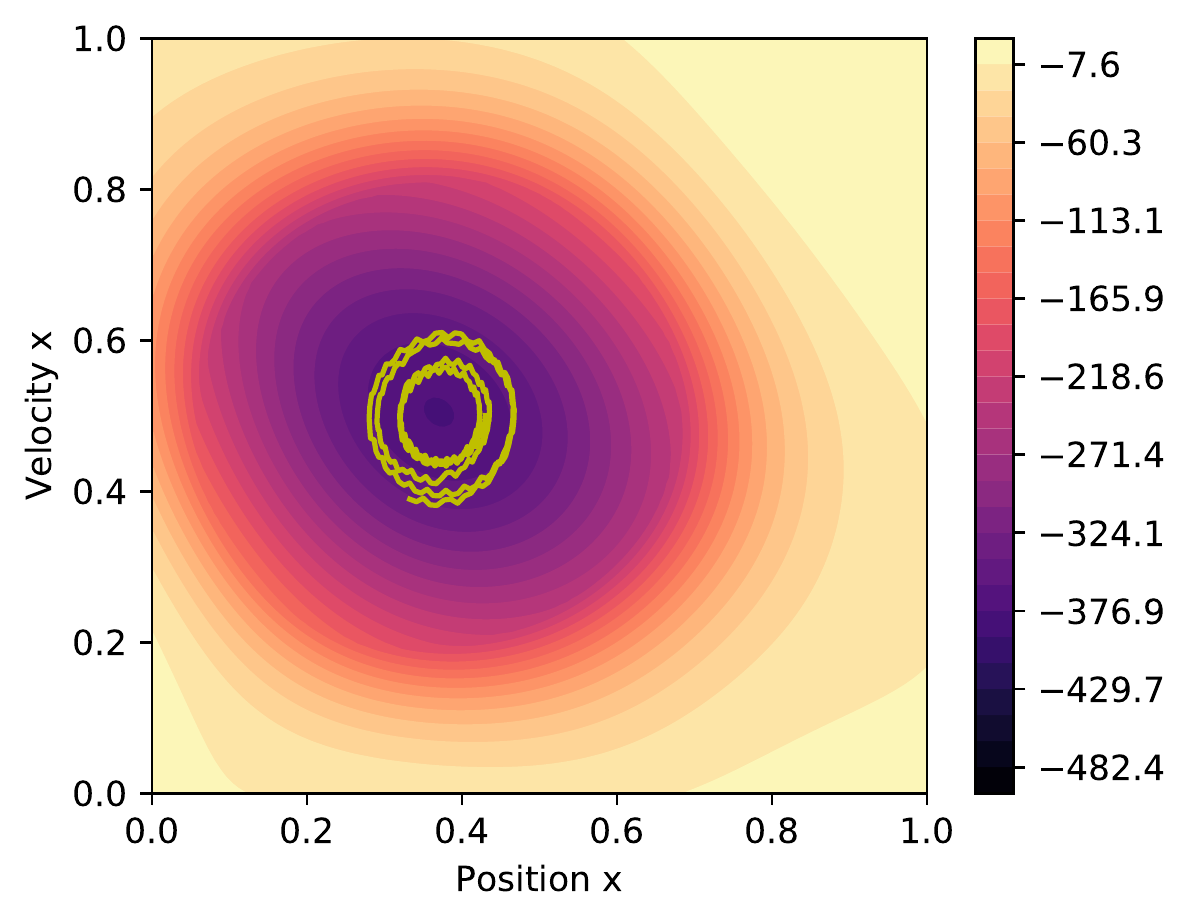}
	}
	\subfloat[][]{
   \includegraphics[width=.23\columnwidth]{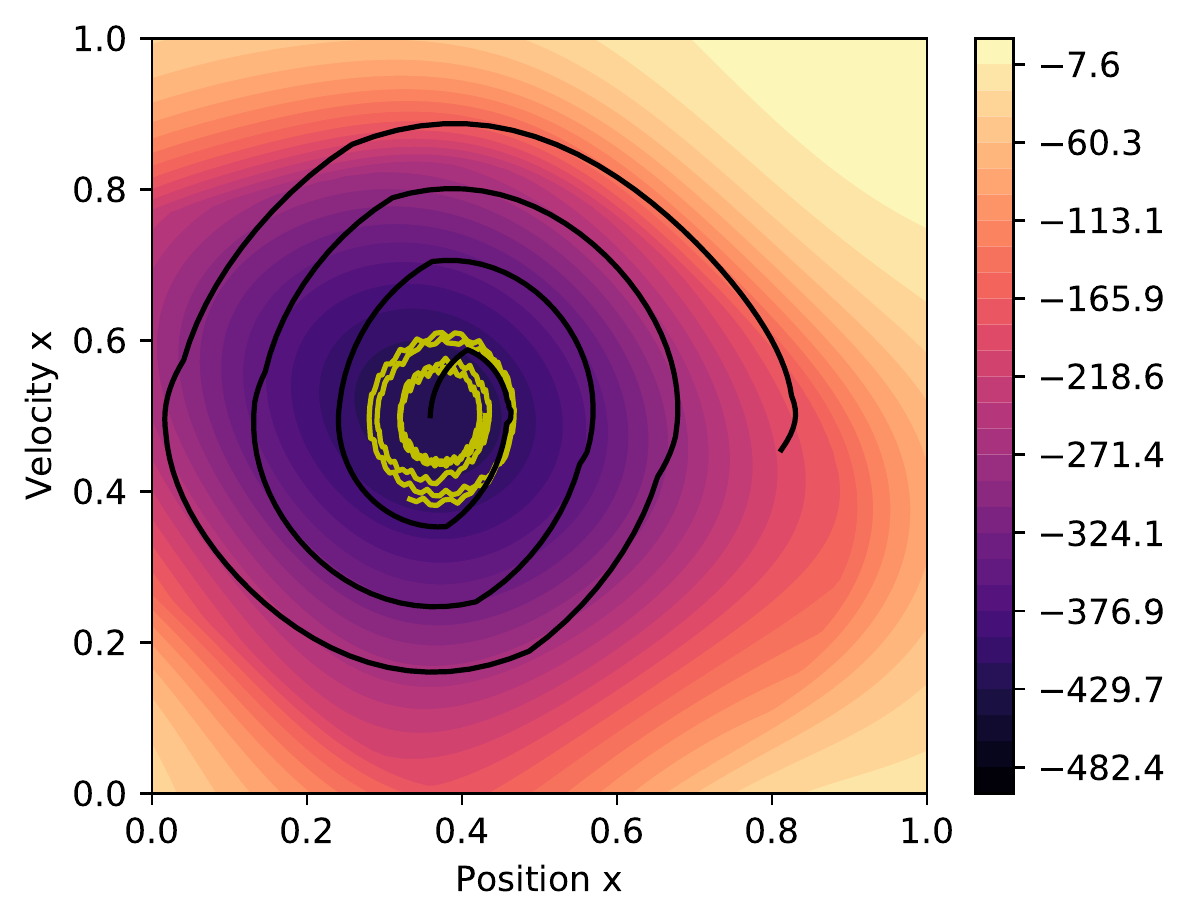}
	}
	\subfloat[][]{
   \includegraphics[width=.23\columnwidth]{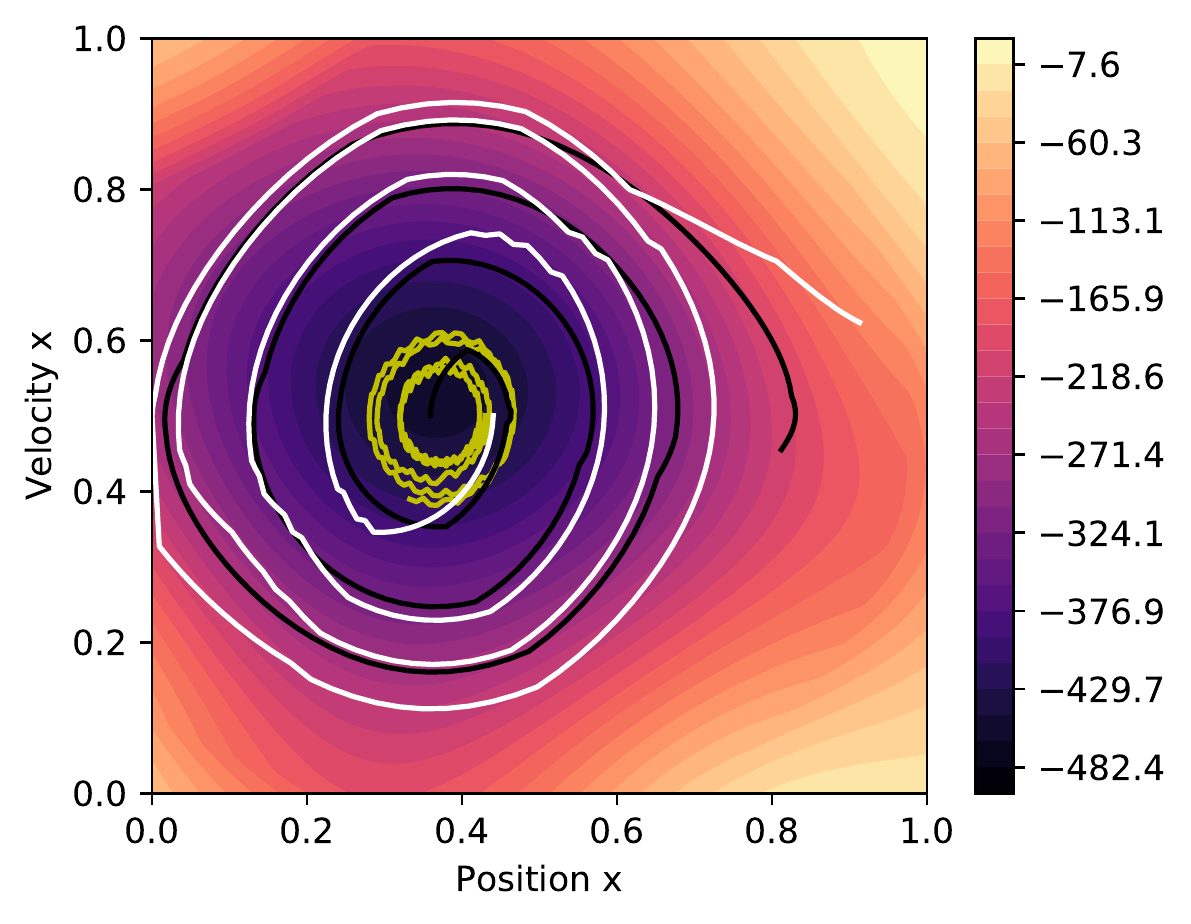}
	}
	\subfloat[][]{
   \includegraphics[width=.23\columnwidth]{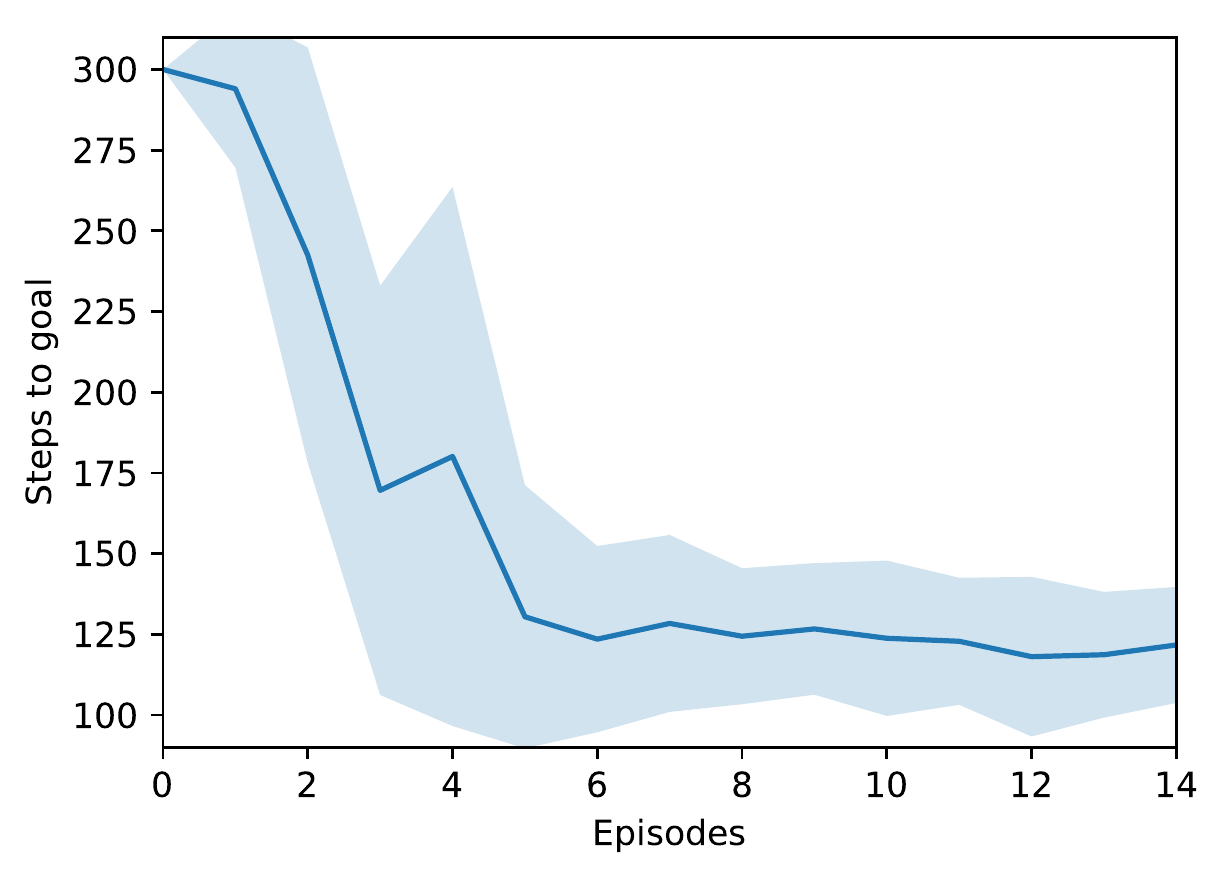}
   \label{fig:mountainCarLearning}
	}
   
	\caption{Goal-only MountainCar. (a,b,c) Exploration value function $U$ (for action $0$) after $1$, $2$, and $3$ episodes. State trajectories for $3$ episodes are plain lines (yellow, black and white respectively). (d) Steps to goal ($x>0.9$), with policy refined after goal state was found (averaged over 30 runs).\label{fig:mountainCarTrajectories}}
\end{figure*}
\subsubsection{Exploration Behaviour on goal-only MountainCar}
We first provide intuition behind what EMU-Q learns and illustrate its typical behaviour on a continuous goal-only version of MountainCar.
In this domain, the agent needs to drive an under-actuated car up a hill by building momentum. The state space consists of car position and velocity, and actions ranging from $-1$ to $1$ describing car wheel torque (absolute value) and direction (sign). The agent is granted a unit reward for reaching the top of the right hill, and zero elsewhere.

Figure~\ref{fig:mountainCarTrajectories} displays the state-action exploration value function $U$ at different stages of learning, overlaid by the state-space trajectories followed during learning. The first episode (yellow line) exemplifies action babbling, and the car does not exit the valley (around $x=0.4$). On the next episode (black line), the agent finds sequences of actions that allow exiting the valley and exploring further areas of the state-action space. Lastly, in episode three (white line), the agent finds the goal ($x>0.9$). This is done by adopting a strategy that quickly leads to unexplored areas, as shown by the increased gap between white lines.
The exploration value function $U$ reflects high uncertainty about unexplored areas (yellow), which shrink as more data is gathered, and low and decreasing uncertainty for often visited areas such as starting states (purple). Function $U$ also features a gradient which can be followed from any state to find new areas of the state-action space to explore.
Figure~\ref{fig:mountainCarLearning} shows  EMU-Q's exploration capabilities enables to find the goal state within one or two episodes.

\subsubsection{Continuous control benchmark}
We now compare our algorithm on the complete benchmark of $7$ continuous control goal-only tasks.
All domains make use of OpenAI Gym \cite{gym}, and are modified to feature goal only rewards and continuous state-action spaces with dimensions detailed in Table \ref{tab:paramValues}.
More specifically, the domains considered are MountainCar and the following:
\begin{itemize}
    \item {\it SinglePendulum}: The agent needs to balance an under-actuated pendulum upwards by controlling a motor's torque a the base of the pendulum. A unit reward is granted when the pole (of angle with vertical $\theta$) is upwards: $\theta < 0.05\text{ }rad$.
    \item {\it DoublePendulum}: Similarly to SinglePendulum, the agent's goal is to balance a double pendulum upwards. Only the base joint can be controlled while the joint between the two segments moves freely. The agent is given a unit reward when the tip of the pendulum is close to the tallest point it can reach: within a distance $d<1$.
    \item {\it CartpoleSwingUp}: This domain features a single pole mounted on a cart. The goal is to balance the pole upwards by controlling the torque of the under-actuated cart's wheels. Driving the cart too far off the centre ($|x|>2.4$) results in episode failure with reward $-1$, and managing to balance the pole ($cos(\theta) > 0.8$, with $\theta$ the pole angle with the vertical axis) yields unit reward. Note that contrary to classic Cartpole, this domain starts with the pole hanging down and episodes terminate when balance is achieved.
    \item {\it LunarLander}: The agent controls a landing pod by applying lateral and vertical thrust, which needs to be landed on a designated platform. A positive unit reward is given for reaching the landing pad within distance $d<0.05$ of its center point, and a negative unit reward is given for crashing or exiting the flying area.
    \item {\it Reacher}: A robotic manipulator composed of two segments can be actuated at each of its two joints to reach a predefined position in a two-dimensional space. Bringing the manipulator tip within a distance $d<0.015$ of a random target results in a unit reward.
    \item {\it Hopper}: This domain features a single leg robot composed of three segments, which needs to propel itself to a predefined height. A unit reward is given for successfully jumping to height $h>1.3$, and a negative unit reward when the leg falls past angle $|\theta| > 0.2$ with the vertical axis.
\end{itemize}

\begin{table}
	\centering
	\begin{tabular}{ l | c c | c c c c c}
		\textbf{Domain} & $d_\mathcal{S}$ & $d_\mathcal{A}$ & $l_\mathcal{S}$ & $l_\mathcal{A}$ & $M$ & $\alpha$ & $\beta$\\
	\hline
	SinglePendulum & 3 & 1 & 0.3 & 0.3 & 300 & 0.001 & 1.0\\
	MountainCar & 2 & 1 & 0.3 & 10 & 300 & 0.1 & 1.0\\
	DoublePendulum & 6 & 1 & 0.3 & 0.3 & 500 & 0.01 & 1.0\\
	CartpoleSwingUp & 4 & 1 & 0.8 & 1.0 & 500 & 0.01 & 1.0\\
	LunarLander & 8 & 2 & 0.5 & 0.3 & 500 & 0.01 & 1.0\\
	Reacher & 11 & 2 & 0.3 & 0.3 & 500 & 0.001 & 1.0\\
	Hopper & 11 & 3 & 0.3 & 0.3 & 500 & 0.01 & 1.0\\
\end{tabular}
\caption{Experimental parameters for all $7$ domains\label{tab:paramValues}}
\end{table}

Most methods in the sparse rewards literature address domains with discrete states and/or action spaces, making it hard to find baselines to compare EMU-Q to. Furthermore, classic exploration techniques such as $\epsilon$-greedy fail on these domains.
We compare our algorithm to three baselines: VIME, DORA and RFF-Q.
VIME \cite{houthooft2016vime} defines exploration as maximizing information gain about the agent's belief of environment dynamics.
DORA \cite{fox2018dora}, which we run on discretized action spaces, extends visitation counts to continuous state spaces.
Both VIME and DORA use additive rewards, as opposed to EMU-Q which uses exploration values.
Q-Learning with $\epsilon$-greedy exploration and RFF is denoted RFF-Q. Because it would fail in domains with goal-only rewards, it is run with classic rewards; see \cite{gym} for details on classic rewards.

We are interested in comparing exploration performance, favouring fast discovery of goal states.
To reflect exploration performance, we measure the number of episodes required before the first positive goal-reward is obtained. This metric reflects how long pure exploration is required for before goal-reaching information can be taken advantage of to refine policies, and hence directly reflects exploration capabilities.
Parameter $\gamma$ is set to $0.99$ for all domains and episodes are capped at $500$ steps. State spaces are normalized, and Random Fourier Features approximating square exponential kernels are used for both state and action spaces with EMU-Q and RFF-Q. The state and action kernel lengthscales are denoted as $l_{\mathcal{S}}$ and $l_{\mathcal{A}}$ respectively. Exploration and exploitation trade-off parameter $\kappa$ is set to $1/\mathbb{V}_{max}$ for all experiments. Other algorithm parameters were manually fixed to reasonable values given in Table \ref{tab:paramValues}.

\begin{table}
	\centering
    \resizebox{\columnwidth}{!}{
	\begin{tabular}{l|c c|c c|c c||c c}
		\multirow{2}{3cm}{\textbf{Domain}} & \multicolumn{2}{c|}{\textbf{EMU-Q}} & \multicolumn{2}{c|}{\textbf{VIME}} & \multicolumn{2}{c||}{\textbf{DORA} (discrete)} & \multicolumn{2}{c}{\textbf{RFF-Q} \small{(classic reward)}}\\
		& \scriptsize{Success} & \scriptsize{Episodes to goal} & \scriptsize{Success} & \scriptsize{Episodes to goal} & \scriptsize{Success} & \scriptsize{Episodes to goal} & \scriptsize{Success} & \scriptsize{Episodes to goal}\\
	\hline
	SinglePendulum  & \textbf{100}\%  & 1.80 (1.07)  & 95\% & 2.05 (2.04)  & 35\% & 3.00 (4.11)   & 100\% & 1.0 (0.00)\\
	MountainCar     & \textbf{100}\% & 2.95 (0.38)   & 65\% & 5.08 (2.43)  & 0\% & --             & 100\% & 8.6 (8.05)\\
	DoublePendulum  & \textbf{100}\% & 1.10 (0.30)   & 90\% & 3.61 (2.75)  & 0\% & --             & 100\% & 4.20 (2.25)\\
	CartpoleSwingup & \textbf{90}\% & 12.40 (16.79)  &  65\% & 3.23 (2.66) & 35\% & 48.71 (30.44) & 100\% & 9.70 (12.39)\\
	LunarLander     & \textbf{100}\% & 28.75 (29.57) &  75\% & 4.47 (2.47) & 30\% & 35.17 (31.38) & 95\% & 19.15 (24.06)\\
	Reacher         & \textbf{100}\% & 19.70 (20.69) & 95\%  & 3.68 (2.03) & 35\% & 1.00 (0.00)   & 95\% & 26.55 (25.58)\\
	Hopper          & \textbf{60}\% & 52.85 (39.32)  & 40\% & 5.62 (3.35)  & 20\% & 30.50 (11.80) & 80\% & 41.15 (35.72)\\
    \end{tabular}
    }
    \caption{Results for all $7$ domains, as success rate of goal finding within 100 episodes and mean (and standard deviation) of number of episodes before goal is found. Success rate rate is more important than number of episodes to goal. Results averaged over 20 runs. DORA was run with discretized actions, and RFF-Q with $\epsilon$-greedy exploration on domains with classic rewards.\label{tab:controlResults}}
\end{table}

Results displayed in Table \ref{tab:controlResults} indicate that EMU-Q is more consistent than VIME or DORA in finding goal states on all domains, illustrating better exploration capabilities. The average number of episodes to reach goal states is computed {\em only} on successful runs. EMU-Q displays better goal finding on lower dimension domains, while VIME tends to find goals faster on domains with higher dimensions but fails in more occasions. Observing similar results between EMU-Q and RFF-Q confirms that EMU-Q can deal with goal-only rewards without sacrificing performance.

\subsection{Jaco Manipulator}
\begin{figure*}
\centering
    \subfloat[][]{
    \includegraphics[width=.23\columnwidth]{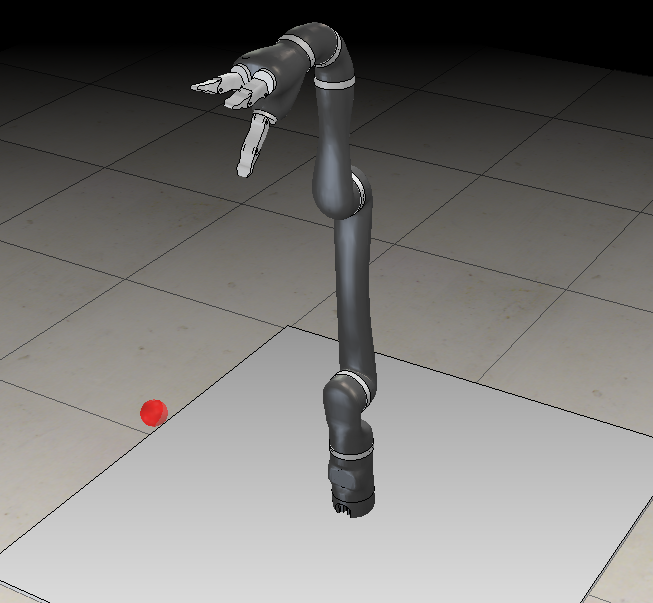}
    \label{fig:jacoImage}
    }
    \subfloat[][]{
    \includegraphics[width=.30\columnwidth]{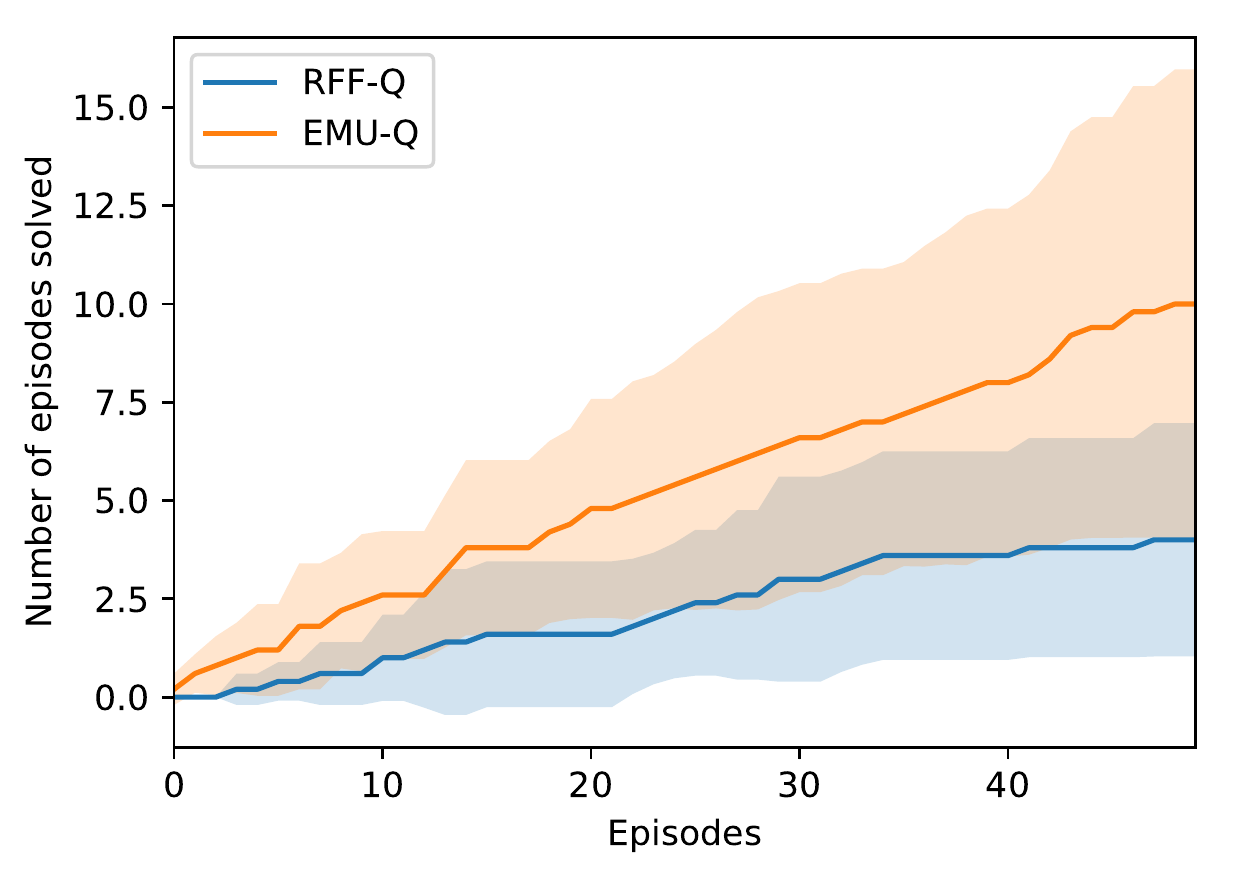}
    \label{fig:jacoResult}
    }
	\caption{(a) Manipulator task: learning to reach a randomly located target (red ball). (b) EMU-Q's directed exploration yields higher performance on this task compared to RFF-Q with $\epsilon$-greedy exploration.}
\end{figure*}
In this final experiment, we show the applicability of EMU-Q to realistic problems by demonstrating its efficacy on an advanced robotics simulator. In this robotics problem, the agent needs to learn to control a Jaco manipulator solely from observing joint configuration. Given a position in the 3D space, the agent's goal is to bring the manipulator finger tips to this goal location by sending torque commands to each of the manipulator joints; see Figure~\ref{fig:jacoImage}.
Designing such target-reaching policies is also known as inverse kinematics for robotic arms, and has been studied extensively. Instead, we focus here on \emph{learning} a mapping from joint configuration to joint torques on a \emph{damaged} manipulator. When a manipulator is damaged, previously computed inverse kinematics are not valid anymore, thus being able to learn a new target-reaching policy is important.

We model damage by immobilizing four of the arm joints, making previous inverse kinematics invalid.
The target position is chosen randomly to form locations across the reachable space. Episodes terminate with unit reward when the target is reached within $50$ steps, zero rewards are given otherwise. We compare EMU-Q and RFF-Q on this domain, both using $600$ random Fourier features approximating an RBF kernel. Parameters $\alpha, \beta$ and $\kappa$ were manually selected to acceptable values of $0.1, 1.0$ and $0.1$ respectively.
Figure~\ref{fig:jacoResult} displays results averaged over $10$ runs. The difference in number of episodes solved shows EMU-Q learns and manages to complete the task more consistently than RFF-Q.
This confirms that directed exploration is beneficial, even in more realistic robotics scenario.


\section{Conclusion}
\label{sec:conclusion}
We proposed a novel framework for exploration in RL domains with very sparse or goal-only rewards. The framework makes use of multi-objective RL to define exploration and exploitation as two key objectives, bringing the balance between the two at a policy level. This formulation has several advantages over traditional exploration methods. It allows direct and online control over exploration, without additional computation or training. Strategies for such control were shown to experimentally outperform classic intrinsic RL on several aspects.
We demonstrated scalability to continuous state-action spaces by presenting EMU-Q, a method based on our framework, guiding exploration towards regions of higher value-function uncertainty.
EMU-Q was experimentally shown to outperform classic exploration techniques and other intrinsic RL methods on a continuous control benchmark and on a robotic manipulator.

As future work, we would like to investigate how exploration as multi-objective RL can be brought to other types of RL methods such as policy gradient. This extension would enable control over exploration in domains with larger state-action spaces, an potentially numerous real-life application.
Other interesting extensions include bringing the online control over exploration achieved by this work to life-long RL, where it would be beneficial. Indeed, exploration can be tuned down in critical situations where high performance is necessary, or increased when learning new behaviours is required.

\appendix
\section{Fourier Basis Features}
\label{app:fourier_basis_features}
Fourier basis features are described in \cite{konidaris2011value} as a linear function approximation based on Fourier series decomposition. Formally, the order-$n$ feature map for state $\bm{s}$ is defined as follows:
\begin{equation}
    \bm{\phi}(\bm{s}) = cos(\pi \bm{s}^T \bm{C}),
\end{equation}
where $C$ is the Cartesian product of all $c_j \in \{0, ..., n\}$ for $j=1,..,d_\mathcal{S}$. Note that Fourier basis features do not scale well. Indeed, the number of features generated is exponential with state space dimension.

While Fourier basis features approximate value functions with periodic basis functions, random Fourier features are designed to approximate a kernel function with similar basis functions. As such, they allow recovering properties of kernel methods in the limit of the number of features. Additionally, random Fourier features scale better with higher dimensions.

\section{Derivation of Bayesian linear regression for Q-learning}
The likelihood function is defined as follows
\begin{equation}
	p(\bm{t}|\bm{x},\bm{w},\beta) = \prod_{i=1}^N \mathcal{N}(t_i|r_i+\gamma\bm{w}^T\bm{\phi}_{s'_i,a'_i}, \beta^{-1}),
\end{equation}
where independent transitions are denoted $x_i=(s_i,a_i,r_i,s'_i,a'_i)$.
we treat the linear regression weights $\bm{w}$ as random variables and introduce a Gaussian prior
\begin{equation}
    p(\bm{w}) = \mathcal{N}(\bm{w}|\bm{0},\alpha^{-1}\bm{I})
\end{equation}
The weight posterior can be computed analytically with Bayes rule, resulting in a normal distribution
\begin{equation}
    p(\bm{w}|\bm{t}) = \frac{p(\bm{t}|\bm{w})p(\bm{w})}{p(\bm{t})} = \mathcal{N}(\bm{w}|\bm{m}_Q,\bm{S})
\end{equation}
Expressions for the mean $\bm{m}_Q$ and variance $\bm{S}$ follow from general results of products of normal distributions \cite{bishop2006pattern}:
\begin{align}
	\bm{m}_Q &= \beta\bm{S}\bm{\Phi}_{\bm{s},\bm{a}}^T(\bm{r} + \gamma \bm{Q}')\\
	\bm{S} &= (\alpha\bm{I}+\beta\bm{\Phi}_{\bm{s},\bm{a}}^T\bm{\Phi}_{\bm{s},\bm{a}})^{-1},
\end{align}
where $\bm{\Phi}_{\bm{s},\bm{a}} = \{\bm{\phi}_{s_i,a_i}\}^N_{i=1}$, $\bm{Q}' = \{Q(s'_i,a'_i)\}^N_{i=1}$, and $\bm{r} = \{r_i\}^N_{i=1}$.
The predictive distribution $p(t|\bm{x},\bm{t}, \alpha, \beta)$ can be obtain by weight marginalization and is also normal
\begin{equation}
    p(t|\bm{x},\bm{t}, \alpha, \beta) = \int p(t|\bm{w}, \beta)p(\bm{w}|\bm{x}, \bm{t}, \alpha, \beta) d\bm{w} = \mathcal{N}(t|Q,\sigma^2)
\end{equation}
Expressions for Q and $\sigma^2$ follow from general results \cite{bishop2006pattern}, yielding
\begin{align}
Q(s,a) = \mathbb{E}[p(t|\bm{x},\bm{t}, \alpha, \beta)] &= \bm{\phi}_{s,a}^T\bm{m}_Q,\\
\sigma^2(s,a) = \mathbb{V}[p(t|\bm{x},\bm{t}, \alpha, \beta)] &= \beta^{-1} + \bm{\phi}_{s,a}^T\bm{S}\bm{\phi}_{s,a}.
\end{align}

\vskip 0.2in
\bibliography{references}
\bibliographystyle{theapa}

\end{document}